\documentclass{article}

\usepackage[format=hang, justification=raggedright, singlelinecheck=false]{caption}
\usepackage{PRIMEarxiv}
\usepackage[utf8]{inputenc}
\usepackage{amsmath}
\usepackage{array}

\usepackage{verbatim}
\usepackage{algorithm}
\usepackage{algpseudocode}

\usepackage{longtable}
\usepackage{tabularx}
\newcolumntype{L}{>{\raggedright\arraybackslash}X}

\usepackage{multirow}
\usepackage{tikz}
\usetikzlibrary{positioning,shapes.geometric,arrows.meta}
\usepackage{booktabs}

\usepackage[utf8]{inputenc}   % allow utf-8 input
\usepackage[T1]{fontenc}    	% use 8-bit T1 fonts
\usepackage{hyperref}              % hyperlinks
\usepackage{url}                         % simple URL typesetting
\usepackage{booktabs}            % professional-quality tables
\usepackage{amsfonts}            % blackboard math symbols
\usepackage{nicefrac}              % compact symbols for 1/2, etc.
\usepackage{microtype}          % microtypography
\usepackage{lipsum}				   % fillers
\usepackage{fancyhdr}            % header
\usepackage{graphicx}            % graphics
\usepackage{appendix}

\title{Continuously evolving rewards in an open-ended environment}

\author{
	Bailey, R.M. \\
	Oxford University Centre for the Environment\\
	Univeresity of Oxford, UK\\
	\texttt{richard.bailey@ouce.ox.ac.uk} \\
}

\begin{document}
	\maketitle
	
	\begin{abstract}
		Unambiguous identification of the rewards driving behaviours of entities operating in complex open-ended real-world environments is difficult, partly because goals and associated behaviours emerge endogenously and are dynamically updated as environments change. Reproducing such dynamics in models would be useful in many domains, particularly where fixed reward functions limit the adaptive capabilities of agents. Simulation experiments described assess a candidate algorithm for the dynamic updating of rewards, RULE: Reward Updating through Learning and Expectation. The approach is tested in a simplified ecosystem-like setting where experiments challenge entities' survival, calling for significant behavioural change. The population of entities successfully demonstrate the abandonment of an initially rewarded but ultimately detrimental behaviour, amplification of beneficial behaviour, and appropriate responses to novel items added to their environment. These adjustment happen through endogenous modification of the entities' underlying reward function, during continuous learning, without external intervention.
	\end{abstract}
	
	
	\section{Introduction and context}
		This paper addresses the difficulty of simulating agents which operate in complex, open-ended environments, where the goals and objective functions of the individuals are incompletely defined. Real-world entities in open environments (e.g. humans, animals) may face multiple concurrent challenges and undertake a broad set of potentially competing tasks. They must learn how to behave without the opportunity of replaying and improving upon past choices, operating on a \textit{linear timeline}. Changes in the environment, and the co-evolution of interactions between multiple individuals, mean the behaviour required for survival may change significantly over time. In the case of biological life, evolution provides opportunities for long-term adaptation, while short-term learning drives behavioural responses, potentially over multiple lifetimes through `cultural inheritance’. This paper is an attempt to capture this rather constrained adaptability within a simulated environment, using embedded Reinforcement Learning (RL) (~\cite{arulkumaran2017deep}) agents and a simplified evolutionary process. 
		In RL, scalar-valued rewards serve as feedback to guide learning, typically generated externally to the agent using human-specified reward functions. Achieving the desired behaviour reduces largely to the problem of defining appropriate reward functions, which ideally provide effective (frequent and unambiguous) training signals. In well-defined tasks with clear goals (‘achieve the highest score in game X’, or ‘place object A at position B’) shaped external reward functions designed by humans can be highly successful (e.g. ~\cite{Mnih2015}, ~\cite{gu2016}, ~\cite{silver2016mastering}, ~\cite{alphaFold}). Taking this further, when the goal of agents can be explicitly stated, proximity of learnt agent behaviour to the desired goal can provide a signal with which to optimise reward functions, further improving performance (~\cite{faust2019}; see also the \textit{Problem statement} below). 
		However, in many complex open-ended environments it is not possible to know the reward functions of relevant entities. An alternative approach to designing or optimizing reward functions is to infer an agent's objectives, values, or rewards from observation of example behaviour, using \textit{Inverse reinforcement learning} ~\cite{AroraAndDoshi2021}. While powerful in many contexts, this approach has a number of significant difficulties in the present case:  it cannot produce reward functions relevant to novel (unseen) situations and faces challenges when example behaviour is ambiguous or suboptimal; it struggles computationally in high-dimensional spaces and can misinterpret actions driven by unseen features; the method can yield multiple plausible reward functions, potentially over-fitting to limited data (~\cite{AroraAndDoshi2021}, ~\cite{Dexter2021}). In the present open-ended context, \textit{a priori} observations are not assumed available, and are of course not possible for novel environmental contexts. 
		If a relevant dense reward function is not known, and can not be estimated, a possible solution is to embody high level goals as the reward. For example in the case of a simulated animal, a large reward for each surviving offspring might encourage multiple required subsidiary behaviours to be learnt. A drawback is the sparse reward signal coming from the environment: multiple (un-rewarded) behaviours are needed, potentially multiple times, before the high level reward can be obtained and it is unlikely these would be undertaken by chance, hence slowing the learning process. More fundamentally, the problem of how to choose the high-level reward to drive the lower level behaviours persists.
		The problem of sparseness can in some cases be addressed through \textit{internal/intrinsic} rewards, which encourage exploration of novel behaviours and system states (~\cite{lopes2012exploration}, ~\cite{poupart2006analytic}, ~\cite{campero2020learning}, ~\cite{karayanni2022extrinsic}) and/or improvements in knowledge of the environment (~\cite{schmidhuber1991possibility}, ~\cite{chentanez2004intrinsically}; ~\cite{stadie2015incentivizing}, ~\cite{houthooft2016vime}). Curiosity-Driven Exploration is one such approach (~\cite{oudeyer2007intrinsic}, ~\cite{oudeyer2007intrinsic}, ~\cite{pathak2017curiosity}, ~\cite{zheng2021episodic}), where rewards reflect how well the agent predicts the outcome of its actions or how surprised it is by its observations. Over repeated periods of learning these approaches can yield impressive results (e.g. ~\cite{pathak2017curiosity}). In`novelty search' (~\cite{risi2009novelty}, ~\cite{Lehman2011}, ~\cite{conti2018improving}), rewards are earned by diverging from stored memories of prior behaviour, and the novelty metric creates a constant pressure to innovate through state exploration. In the present case, a drawback of these approaches is they are not guaranteed to consistently find and exploit specific objectives, particularly for more complex problems. Behaviours which are `good' in some sense for agents (e.g. in aiding survival) may be abandoned for more novel behaviours.
		At a deeper level, it remains the case that prior decisions are made which to a large extent direct the behaviour of the agents, here deciding time should be spent exploring above all else, which may not reflect the behaviour of real-world agents in linear timelines. Additionally, human-chosen extrinsic rewards may still be required in order to achieve specific goals (e.g. producing offspring) as a drive to explore will alone not necessarily lead to discovery of behaviours unrelated to the agent's own ability to explore.
		In the realm of evolutionary algorithms, rewards directly inspired by the survival challenges of living entities can be operationalized through an explicit (modeller-defined) fitness function which promotes selection for progressively fitter generations (e.g. ~\cite{moriarty1999evolutionary}, ~\cite{lambora2019genetic}, ~\cite{faust2019}, ~\cite{Hallawa_et_al_2020}). In the present work importance is given to avoiding the modeller setting permanent objectives (defining \textit{success}), as this is ultimately responsible for directing behaviour, and therefore not necessarily adaptive to novel environmental conditions. An alternative to this is to use survival itself to implicitly define success (only those which behave `well' will survive).  This essentially calls for adaptive agents which determine their own goals, transcribe these in to dense external rewards, and promote the learning of beneficial behaviours on a linear timescale. A complete approach would also generalise beyond adjusting rewards for existing behaviours, to including means of discovering and rewarding new actions in appropriate ways. 

		%Consciousness is the solution to the problem of solving general problems which algorithm design cannot foresee...
		
		\paragraph{This contribution}  
		Through simulation, this paper shows endogenous updating of a dense external reward function can produce simulated populations of agents which dynamically adjust to challenges within an open-ended environment: a self-replenishing population of agents continuously learn and dynamically adapt their rewards (and therefore their behaviour) as time proceeds. This is in contrast to existing RL approaches that attempt to find optimal policies for fixed reward functions. While agent behaviour can adapt to variations in its environment during standard RL training (and potentially in subsequent inference), because the reward function remains fixed, behavioural adaptation is essentially finding different solutions to meet the same objective, rather than reformulating the objectives (through the reward function) to meet the requirements of new challenges.
		An abstract ecosystem-like simulated environment is used to test these ideas: entities with finite lifetimes possess initial reward functions comprising multiple components; entities inherit (from their maternal parent) the value for each reward coefficient and also an expectation over time of rewards for each component; when they become parents they pass on an updated version of both the reward coefficient and its expectation; the modification is based on their own experience. This process is termed RULE: Reward Updating through Learning and Expectation (~\cite{bailey2023}). This paper presents a conceptually simple test of this approach in which detrimental/advantageous options are made available to simulated entities, who adjust their reward functions in ways which are advantageous to them in avoiding extinction, without being explicitly encouraged to survive.
		
	\subsection{Scope and Problem statement}
		The context of this work is the behaviour of entities, and populations of entities, operating in open world contexts, with constraints on, for example, information availability, interaction rates, energy use/availability, and transfer of currencies (in the broad sense), which affect the ability of entities to survive. The environment is open in the sense there is no clear task to complete, and there are multiple simultaneous influences on entities, with the possibility of new phenomena at any time. These features are seen in many real-world contexts including ecological-, economic-, and social-systems and the generalisability of the current findings is discussed in later sections. An abstract ecosystem-like model is used here as a test case, which hosts a single type/species of agent entity (named `Ents') and a single type of primary producer (`PP').
		
		\paragraph{Problem statement}  The true state of the world is only partially observable by each Ent agent in the simulation. The learning problem conforms to a Partially-Observable Markov Decision Process, described by the tuple $\langle S,A,O,D,R,\gamma \rangle$, where $S=\left\{s_1,s_2,\ldots,s_n\right\}$ is a set of partially observable states, $A=\left\{a_1,a_2,\ldots,a_m\right\}$ is a set of actions, and $O=\left\{o_1,o_2,\ldots,o_k\right\}$ is a set of observations. 
		$\boldsymbol{D} : S\times A\times S \rightarrow [0,1]$  models the unknown system dynamics as a transition distribution.
		The immediate reward, $R(\theta, s, a) : S \times A \rightarrow \mathbb{R}$, where $\boldsymbol{\theta}\in\Theta$ are the reward coefficients, is then discounted at a rate $\gamma\in[0,1]$. 
		Reinforcement Learning (RL) is an attempt to find a policy $\pi_R^\ast$ that maximizes the expected cumulative discounted reward $R$ of trajectories $\mathcal{T}$ drawn from a set of initial conditions $N\subset O$, such that 
		$\pi_R^\ast=\arg\max_{\pi}\mathbb{E}_{\mathcal{T}(N,\boldsymbol{D},\boldsymbol\pi)}\left[\sum_{i=0}^{\left|\mathcal{T}\right|}{\gamma^iR\left({\boldsymbol{\theta},\boldsymbol{s}}_i,\boldsymbol{a}_i\right)}\right]$. 
		The behaviour of the agent is therefore dependent on the reward function $R$, which is typically shaped through human selection of $\boldsymbol{\theta}$ to result in the desired outcomes. 
		The success of any policy can be quantified as
		$\mathcal{J}\left(\pi\right)=\mathbb{E}_{\mathcal{T}(N,\boldsymbol{D},\boldsymbol\pi)}\left[\mathcal{G}\left(\boldsymbol{\mathcal{T}}\right)\right]$, 
		where $\mathcal{G}$ is the metric for trajectory success.
		In unambiguous RL tasks, the relationship between $R$ and candidate (human-created) evaluation functions $\mathcal{G}$ can be straightforward, and the chosen $\mathcal{G}$ can even be used as a fitness metric against which to evolve successful reward functions (~\cite{faust2019}). Defining success for an agent (or population of agents) in real-world open-ended systems is not so straightforward, as multiple related tasks are underway simultaneously and both the properties of agents and the challenges they face may change over time. 
		Further, the agent may face both behavioural and evolutionary trade-offs which are not explicitly known, and maximizing lower-level rewards may have impacts on achieving higher-level success. For example, maximizing lower-level rewards for consumption might produce longer-lived entities and therefore yield larger more successful populations (higher-level success) even though population size/persistence is not part of the reward definition. Such high-level success could in principle be measured by a suitable metric, but finding a unique such function ($\tilde{G}$) is not expected to be possible in real-world open-ended systems (e.g. finding a unique goal and therefore suitable success metric for an ecosystem). 
		The present work is not therefore an attempt to find an optimally-shaped reward function (i.e. $\arg  {\max_{\theta}}{\mathcal{J}\left(\hat{\pi}\right)}$) 
		that maximizes an explicit success metric. Instead, the goal is to produce an emergent process which endogenously updates $\boldsymbol{\theta}$ in order to maximise the unobservable success metric ($\tilde{G}$). Ideally, the process must explore and adjust both the reward coefficients ($\boldsymbol{\theta}$) and the basic properties of the agents ($\boldsymbol{P}$) to prevailing conditions. 
		Given that HL objectives emerge dynamically from contingent (survival) challenges faced by populations of entities, it is not guaranteed that endogenous processes (through selection pressures) acting on $\boldsymbol{\theta}$ will result in convergence to optimal solutions,
		$\tilde{\pi}=\arg{\max_{\boldsymbol{\theta}}{\tilde{\mathcal{J}}\left(\tilde{\pi}\right)}}$, where
		$\tilde{\mathcal{J}}\left(\tilde{\pi}\right)=\mathbb{E}_{\mathcal{T}(N,\boldsymbol{D},\tilde{\pi})}\left[\tilde{\mathcal{G}}\left(\boldsymbol{\mathcal{T}}\right)\right]$. 
		Indeed, in real-world complex adaptive systems there is ample evidence of entities occupying local physiological and behavioural minima, for example due to inescapable physiological inheritance  (e.g. ~\cite{wedel2011}) or behavioural path-dependence (e.g. ~\cite{Heino2023HumanAttractors}).
	
	\section{Simulated open-world environment}
	In the simulated environment, discrete individual entities (`Ents') with finite lifetimes operate within a spatially extensive physically-based world. The environment enforces certain constraints, including physical constraints on movement, information transfer, and conservation of energy. 
	An external source provides energy in the form of light which powers the system, and reproducing `primary producers' directly use this light to create biomass. All actions have an energy cost and are therefore limited by energy availability. Mobile entities ('Ents') with finite lifetimes exist in this environment, obtaining energy through consumption and have the potential to reproduce. They possess multiple senses and (using reinforcement-learning, RL) learn to map their observations to a fixed set of available actions which include movement, manipulation of objects, emission of signals, and the transferring of various currencies to other entities. Some of the state variables describing each entity are fixed throughout each lifetime and inherited during reproduction through `genes' from the two required parent entities. The demands of the environment naturally produce selection pressures, and these result in evolution.
	The following sections provide brief descriptions of the components of the model. Additional details are given in Appendix A.
	
\begin{figure}[h!]
		\begin{center}
			
			\includegraphics[width=6.3in,height=2.88in, angle=0]{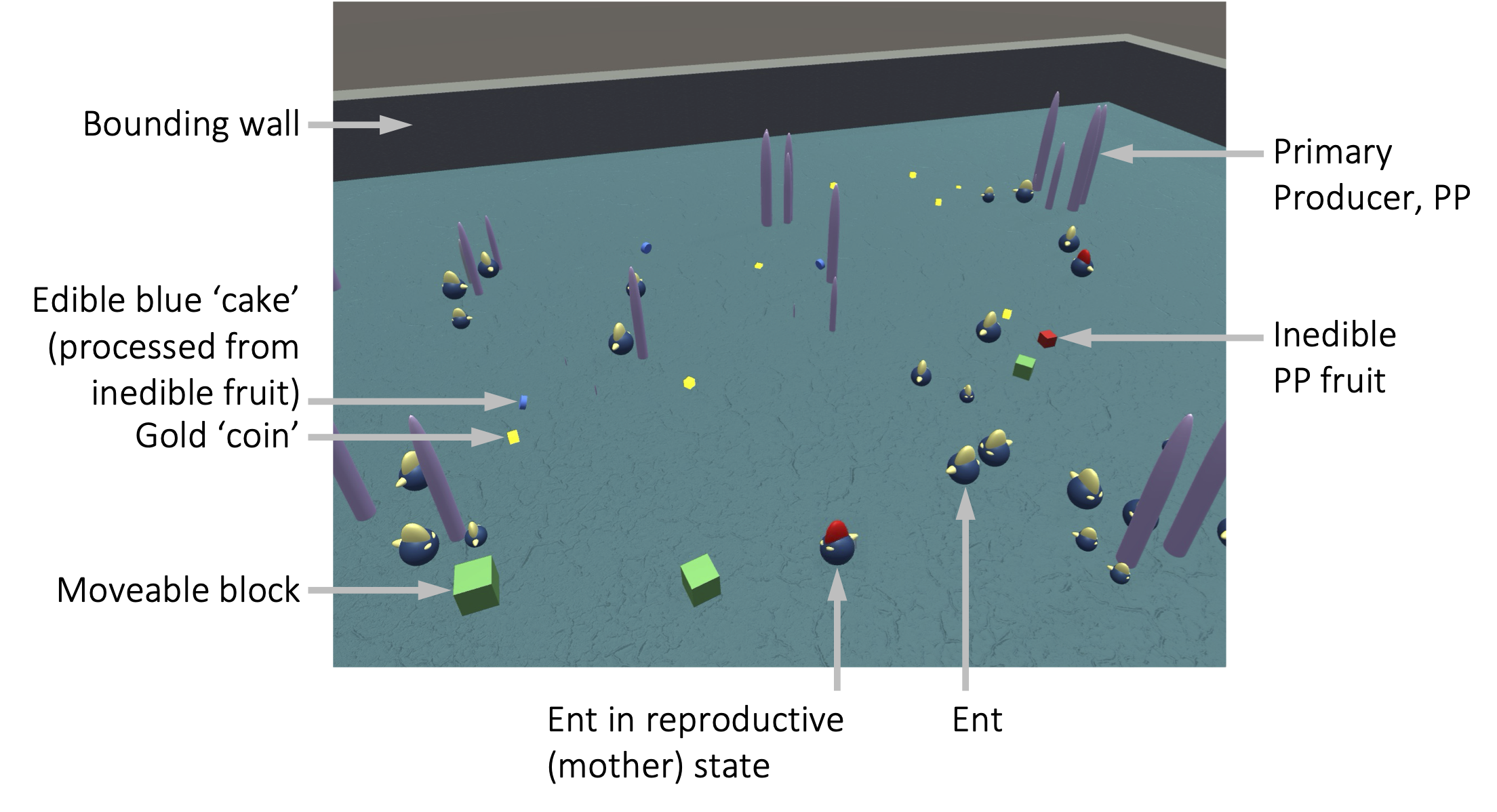}
			%\vskip 0.25cm
			\caption{An image of the model running, showing the environment components. }
			\label{Fig_env_components}
		\end{center}
	\end{figure}

	\paragraph{Simulated physical environment} The simulated environment is a continuous 3D space, with a basal surface area of 20 x 20 m$^2$ and unbounded in the vertical dimension. Time proceeds discretely in integer steps $\Delta t=0.02 s$. Basic physics (gravity, inertia, response to force, drag and scattering during collisions) is applied to all rigid objects (the simulation was built using the Unity 3D game engine, with physics updates implemented using Nvidia PhysX). Time is defined in seconds, mass in kg and length in m. Energy is normalised to the power of the external source, $S$, which remains fixed at 1 throughout all of the reported experiments. The main physical elements of the world are shown in Figure \ref{Fig_env_components}.
	
	\subsection{Primary producers (PP)}
	Primary producers are the base of the food chain and produce biomass (bodymass and fruits) using energy derived directly from the external energy (light) source. 
		
	Information contained in their genes entirely determines their physiology and behavioural responses. PPs grow from seeds ejected within fruits by a parent PP. On the ground, fruits notionally decompose and their seeds lay dormant for a genetically defined period before starting to grow. While on the ground, some fraction of the non-palatable fruits are available and may be collected by Ents (and potentially converted later to \textit{processed food}). Once a seed begins to grow, it grows at a rate proportional to its mass and the strength of the incident light, until it reaches a genetically-set maximum mass. 
	While alive, PPs compete with each other for space, within a competition radius proportional to their height, with the larger of the two killing the smaller. PPs can also be consumed directly by Ents, who remove bite-sized amounts of mass. The palatability of primary producers to Ents (and hence the sign and proportion of the biomass-derived energy available to the Ent) is determined by the genes of both Ent and PP. Eventually, PPs die for one of three reasons: (i) reaching their genetically-determined maximum lifespan; (ii) loosing all mass to consumption; or (iii) being outcompeted for space by a larger nearby PP.
	
	\subsection{Ents}
	Simulated entities ('Ents') inhabit the surface of the environment and are essentially spherical in shape and have a density equivalent to that of water. To aid visual interpretation, additional body parts (eyes, crests, and tails) are given to Ents but these serve no functional purpose. The internal physiological workings of Ents are not resolved, and the ability to perform essential physical operations, such as movement, digestion, growth and reproduction, are assumed. Ents can be damaged through impact, predation or mass loss due to starvation and can therefore die before their maximum potential lifetime is reached. Ent behaviour is learnt collectively using reinforcement learning.
	
	\paragraph{Energy}
	Energy costs are associated with various activities, including the base running cost (resting metabolic rate, proportional to $\text{mass}^{2/3}$), movement (proportional to mass), growth of biomass, synthesizing of currencies, processing raw material (from inedible fruits to processed blue `cakes'), currency transactions, and collision damage (which is proportional to the strength of the impact above a minimum threshold). Ents obtain energy through the consumption of primary producers (PPs), other Ents, and processed `cakes'. 
	To consume food, Ents must physically touch the object and \textit{bite-off} an amount of mass in proportion to their own body mass. This ingested material is then notionally stored within the Ent's stomach and gradually digested to obtain energy. The energy release occurs incrementally over a period that is proportional to the mass consumed. Further consumption is only possible once the stomach has emptied sufficiently. The energy obtained from consumption depends on the energy density of the material and its palatability to the Ent. 
	Palatability is determined by the genetic makeup of both the Ent and material consumed (the palatability of a specific potential foodstuff is therefore subject to evolutionary adaptation); three values within the genes of Ents and PPs determine their \textit{tissue} makeup, and are also translated to RGB to render the Ent/PP colours in visualisations. 
	Energy obtained by digesting consumed material can be stored (as `fat mass') for later use in powering physiological functions, and meeting other costs, or it can go to growing biomass (both for the Ent and any offspring). If an Ent is not fully grown, and it's stored energy level is below a genetically-set amount, energy from digestion is partitioned between growth and storage according to a genetically determined ratio; if the energy store has reached the minimum amount (and the Ent is not fully grown), all energy is used for biomass growth; finally, if the Ent is fully grown, all energy sent to the store for later use. In the absence of available food, Ents automatically draw energy from their energy stores. If no food is available and the energy (fat) store is depleted, the Ent's remaining (lean) body mass is internally converted into energy until it reaches a minimum viable body mass (relative to its maximum previously-achieved mass). Once the minimum viable body mass is reached, the Ent dies.
	
	\paragraph{Currencies}
	By definition, interactions between entities in real systems involve a transfer of some combination of information, material and/or service (loosely termed), and to augment potentially meaningful interactions between Ents, a generalised representation of currencies is introduced in to the simulation. Here, a currency is something that one entity can choose to give to another, and is not necessarily tangible. Two two key properties are used to differentiate four types of currencies: `subtractability' (\textit{True}: currencies are used-up when given; \textit{False}: the currency is not used-up and so forever available) and `synthesizability' (\textit{True}: currencies can and must first be created by individuals, with some material/energy cost; \textit{False}: the currency can only be exchanged once collected). Exchange of currencies may incur a transaction cost, and rewards are associated with their transfer. One of each currency type is used in the current experiments, summarised in Table \ref{tab:currencyTable}.
	
	\begin{table}[h!]
		%\begin{tabularx}{lllp{8cm}}
		\begin{tabularx}{\textwidth}{lllX}
			\toprule
			Currency & Subtractable & Synthesizable & Model component\\
			\midrule
			$C_1$ & True & True & An item requiring energy + raw material to produce. Here, inedible fruits from PPs can be collected and processed by Ents, producing processed material (`cakes') with maximum possible nutritional value, as currency $C_1$  \\
			$C_2$ & False & False & A service-/action-related currency, which is innate and entails no additional cost (grooming, and the display of positive emotions, are examples).  \\
			$C_3$ & False & True & A non-innate skill which can be obtained at some production cost, but is then infinitely reusable (a skill such as the ability to play music is an example). \\
			$C_4$ & True & False & Found objects, typically finite. Here, `coins' are dropped in to the environment periodically, and can be picked-up, dropped again, or given to other Ents; uncollected coins vanish after a set duration. \\
			\bottomrule
		\end{tabularx}
		\caption{Properties of transferable currencies}
		\label{tab:currencyTable}
	\end{table}
	
	\paragraph{Life-cycle, reproduction and inheritance} Each Ent has a finite lifetime, during which it can be both a `father' and `mother' to offspring. Reproduction is not a specific single action in its own right, rather reproduction happens as two Ents touch when either/both of them are in a reproductive (`mother') state. Each Ent possesses an inherited set numerical values that determine its core properties, referred to as its gene, $G=\left\{g_{i},\ldots g_{n_g}\right\}$ (and noting this gross simplification compared to biological genes). The reproductive (`mother') state can occur up to a genetically-determined number of times over each lifetime if/when the Ent reaches a minimum relative size and accumulates sufficient stored energy. Ents can (and do) learn to recognise when others are in this reproductive state. If `Ent-A' (\textit{father}) touches a reproductive `Ent-B' (\textit{mother}), there is an exchange of genetic material ($A \rightarrow B$) and `Ent-B' produces a genetically-mixed offspring (and burdens the associated energy cost of biomass production) (genetic noise is also added during reproduction). The `decision' to mate is therefore expressed through movement. Offspring are born immediately on mating and possess an energy reserve (genetically set) donated by the mother sufficient under initial gene values to keep them alive for \textasciitilde10 s (\textasciitilde5\% of their maximum lifetime) without additional consumption. Offspring possess no other currencies or items, and these must be acquired during their lives. Ents finally die due to either (i) reaching their (genetically-determined) maximum lifespan, or (ii) loss of mass due to predation or starvation; their biomass is removed instantly from the system, along with their stored currencies and any collected items (with the exception of coins, currency $C_4$ ) with persist for 1 $s$ before disappearing).
	
	\subsection{Reinforcement Learning (RL)}
	The ability of Ents to learn (continuously) is an important feature of the present work, and Ent behaviour (the choice of action taken) is learnt from experience using deep RL, as outlined in the introductory section. In some of the experiments described below, learning is continuous meaning decisions are part of the learning process and any change in reward function has potential to feed through to changes in behaviour. In other experiments, a previously learnt policy is used (in inference mode) to decide actions, without being updated. A high level summary of the main component of the process are provided below (technical details of the model and RL process are given in the Appendix).
	
	\paragraph{Observations}
	Each 0.02 \textit{s} time step, entities make a range of observations covering their own internal state and their detectable environment (including other entities). 
	\textbf{`Movement':} horizontal movement (x and z velocity) and rotation angle; 
	\textbf{`Vision':} Ray sensors provide information about visible object type, direction relative to the forward direction of the Ent, and distance (17 rays, equally spaced over ±40 degrees of the forward direction, with a range dependent upon the illumination level, up to a maximum range of 50 m).
	\textbf{`Taste' / palatability:} when Ents consume material, they sense its palatability (based on the paired sums of the three `tissue' genes described above).
	\textbf{`Touch':} Ents can sense the impact damage when they hit any object through an energy cost which is proportional to the impact force (notionally associated to the repair cost).
	\textbf{`Hearing':} Within a genetically-set radius, Ents can identify signals (equal to $\{0, 1, 2\}$), and perceive their direction and distance ($s=0$ is also the sound observation made in the absence of any Ent signals). No intrinsic meaning is given to these signals and Ents are free to use them, or not, and to make their own associations.
	\textbf{`Smell/scent':} Ents can detect the `scent' of the closest Ent within their sensing radius. The `scent' is a single value, calculated as the mean absolute difference between each of three `tissue genes' described above.
	\textbf{`Nearby group size':} the (normalized) number of Ents within sensing radius; 
	\textbf{`Inventories':} the amount of each Currency stored (including the amount of stored energy, $C_0$, and numbers of coins, $C_4$), the `starvation level' (proximity to minimum viable bodymass), amount of stored raw material (PP fruits) and processed material (`blue cakes'), and numbers of cubes held (0 or 1).
	\textbf{Physiological state}: whether ready to mate and give birth (indicated by a red crest); whether `feeling pain' (from being eaten, or from dying before the genetically-determined maximum lifespan); 
	\textbf{Light level:} the intensity of the external light source.
	
	\paragraph{Actions}  The `physiological' processes of Ents (digestion, energy partitioning, growth, gestation) are regularly updated automatically without any direct control by the Ent. Ents can however choose to interact with the environment through various discrete actions (each of which entails an energy cost): moving, picking-up/dropping objects, exchanging currencies and signals, eating, processing raw materials and reproducing. Any set of non-exclusionary actions can be taken in each successive 0.02 \textit{s} time increment, as listed here (details in Appendix A). Those actions which are not logically possible are `masked' (made unavailable) to reduce the action space for learning. For example, the action to drop an object is masked if the object is not currently held. 
	\textbf{Move:} Ents can choose to rotate in $1^\circ$ increments around their vertical axis, $\Delta Rotation = \{0, 1^\circ, -1^\circ\}$; a forward force $\Delta F=\{0,\delta F,-\delta F\}$ can be applied, where the increment $\delta F$ is a function of Ent lean mass, while the force acts on the total body mass.
	\textbf{Eat/bite:} $\{true,false\}$ if Ents are touching a `soft' object they can bite it and take some of its mass. If the palatability score is positive, the material is digested over a period of time (during which no further consumption is possible) and energy is obtained (depending on the energy density and palatability of the material); if palatability is negative, no energy is obtained from digestions and a `poisoning' energy reduction is applied, in proportion to the palatability score.
	\textbf{Emit signal:} Ents choose whether to remain silent (emit signal $0$), or to emit signal $\{1,2\}$. 
	\textbf{Give currency:} When touching another Ent, any currency possessed can be given to that Ent, in increments of $C_{igiven}$, which is determined genetically (G[66-69]) for each currency (except for currency $C_4$ (coins) which is given one-at-a-time); the receiving Ent increases its currency inventory by this amount; the giving Ent reduces its inventory by thid amount only if the currency is `subtractable', and by zero if the currency is `non-subtractable'.
	\textbf{Pick up object:} When touching any movable object with mass less than the Ent, a $\{true,false\}$ choice is made to pick it up; all objects other than blocks can be stored in the Ent's inventory; blocks are carried aloft by the Ent and in this case no other objects can be picked up until the block is dropped.
	\textbf{Drop object:} any object being carried (a cube, coins ($C_4$), fruits or processed `cake' material) can be dropped, with a $\{true,false\}$ choice.
	\textbf{Process raw material in inventory:} If the Ent is carrying a PP seed, the mass of the seed can be converted to a processed `blue cake', which is edible and has the maximum possible palatability (hence, maximum possible energy yield on consumption/digestion). Once processed, the blue-cake must be dropped to be eaten. It can be dropped and picked up to store again as required.
	 
	\paragraph{Rewards and Reward updating}
	In order to create a population of Ents which can potentially go on to update their reward functions endogenously, an initial reward function is required to begin the process of learning (an alternative, not chosen, being to slowly evolve such a function from scratch). Approaches to creating dynamic reward functions are discussed in Section \ref{Discussion}. The chosen reward function comprises a set of minimum essential components (for eating, reproduction and pain) plus components for each of the ($N_C=4$) currencies, to provide potential for richer Ent interactions and for experimental manipulations. The reward obtained at time step $t$ is calculated as
	\begin{equation}
		R = f\theta_{\text{consumption}} + b\theta_{\text{reproduction}} + p\theta_{\text{pain}} \\
		+  \sum\nolimits_{j=1}^{N_C} \Delta C_j \theta_{j}
	\end{equation}
	For each of the $7$ reward components there is a coefficient $\theta$ which scales the contribution to the total. The first term accounts for consumption (of food material) where $f$ is the food mass consumed, the second is reproduction where $b$ is the number of offspring produced (rewards are given to both parent Ents). The third term relates to pain (due to impact damage, being bitten/eaten or dying before the maximum lifespan is reached) and is the only reward coefficient which remains fixed, providing an anchor to which all other reward coefficients are relative. The final term accounts for net changes in currencies. 
	The proposed algorithm for endogenous reward updating is outlined in Algorithm 1 below and it results in updates to $\theta_i$ (where $i$ indexes all components except \textit{pain}). The algorithm can be qualitatively described as follows: each Ent inherits (from its mother) the value of each reward function coefficient $\theta_i$ and also an expectation distribution (over age) of the accumulated reward for each component. If ant Ent becomes a mother, it passes on updated versions of each reward coefficient and each expectation distribution, where the modification depends on its own experience: for each reward component, if the mother Ent has accumulated more reward than expected (within a 20 s age bin), a larger reward value and increased expectation is passed on, and \textit{vice versa}. Ents pass on reward coefficients to their offspring which encourage behaviour closer to their own than had no modifications been made. Ents which are \textit{more successful} in some way, because of their experiences in the environment, will be more likely to pass on their encouragements, and what is effectively a majority vote on reward updates (and therefore eventually behaviour) occurs at the population level.
	
	\textbf{Definitions:}
	$\boldsymbol{\theta}$: the set of reward coefficients; 
	$\boldsymbol{R}$: the set of reward values obtained at time $t$; 
	$\boldsymbol{\alpha}$: the set of reward expectation increments; 
	$\boldsymbol{\beta}$: the set of reward coefficient increments; 
	${E}({\boldsymbol{\tau}})$: the set of initial Expected Accumulated Rewards distributions, across age classes $\boldsymbol{\tau}$
	\begin{algorithm}
		\caption{RULE: Reward Updating through Learning and Expectation}
		\begin{algorithmic}[1] 
			\Require $\boldsymbol{\theta}$, $\boldsymbol{E}(\boldsymbol{\tau})$,  $\boldsymbol{\alpha}$, $\boldsymbol{\beta}$, $\text{lifeStatus} \equiv alive$
			
			\Ensure Offspring with updated Reward Expectations $E_i(t)$ and Reward coefficients $\theta_i$, if offspring produced
			
			\State Initialize $A_i \leftarrow 0, \forall{i}$ \Comment{$i\in{\left[ 1,\mathrm{numberOfRewards}\right] }$}
			\While{$alive$}
			\State $A_i \leftarrow A_i + R_{i,t}, \forall{i}$ \Comment{Incrementally sum rewards each time step $t$}
			\If{producing offspring}
			\State $\tau = f(T)$ \Comment{Calculate age bin $\tau$ based on current age of parent, $T$}
			\State $\Delta_i \leftarrow A_i - E_i(\tau), \forall{i}$  \Comment{Compare accumulated rewards to expected rewards}
			\If{$\Delta_i > 0$}
			\State $\delta_E = \alpha_i$
			\State $\delta_{\theta} = \beta_i$
			\ElsIf{$\Delta_i < 0$}
			\State $\delta_E = -\alpha_i$
			\State $\delta_{\theta} = -\beta_i$
			\Else
			\State $\delta_E = 0$
			\State $\delta_{\theta} = 0$
			\EndIf
			\State $E_i(\tau) \leftarrow E_i(\tau) + \delta_E$ \Comment{Updated Reward Expectations}
			\State $\theta_i \leftarrow \theta_i + \delta_{\theta}$ \Comment{Updated Reward Coefficients}
			\State $\text{Offspring}.E_i(\tau) \leftarrow \text{Parent}.E_i(\tau), \forall \{i,\tau\}$ \Comment{Pass on updated values to offspring}
			\State $ \text{{Offspring}}.\theta_i \leftarrow \text{{Parent}}.\theta_i$, $\forall{i}$
			\EndIf
			\EndWhile
		\end{algorithmic}
	\end{algorithm}
	
	The time complexity of the RULE algorithm is $O(n)$, where $n$ is the number of iterations of the \textit{while} loop (determined by the lifespan of the Ent, of order 10e4-10e5, multiplied by the fixed number of reward coefficients).
	
\section{Experiments}
\subsection{Purpose and overview}
 To test the RULE algorithm assessments were made of whether Ent populations could make significant changes (autonomously) to their reward coefficients (and hence their behaviour, under continuous learning) in order to improve their chances of survival, in response to changing environmental conditions. In Experiments 1 \& 2 outlined below, a challenge to population survival is set up which can only be overcome if the population `chooses' to abandon certain behaviour, running contrary to the initial reward structure (the population must act `against the will' of the initial population). 
The environment of the Ents is set up initially with constant benign conditions and initial reward coefficients ($\boldsymbol{\theta}$) hand-chosen to yield learnt behaviour conducive to a persistent (multi-generational) population. Within this setting the collection of coins (currency C4 ) is rewarded. Collecting coins is an intentionally frivolous activity, providing no immediate benefit and entailing a small  time/energy (therefore opportunity) cost. During Experiment 2, coin availability is progressively increased to the point where so much time/effort could be spent on coin collection that the population would suffer (through starvation and a lack of reproduction). This is the \textit{coins challenge} which the Ent population faces. Experiment 3 follows on from Experiment 2, and the properties of coins are changed such that they affect stored energy and the probability of reproduction. The purpose of this experiment is to assess the ability of Ents to adjust their behaviours when collecting `coins' which provide more direct signals compared to the more implicit opportunity-cost signal associated with coin collection in Experiment 2. Where there are positive benefits, coins are referred to as `vitamins', and when the effects are negative, `poisons'. For the RULE algorithm to be judged successful, the population must survive the \textit{coins challenge} and also be able to modify behaviour in response to the opportunities and risks associated with vitamins and poisons.
The purpose of the final experiment (Experiment 4) was to assess whether reward updating could `switch on' behaviour associated with a previously unseen actions which had (neutral) rewards of zero. This experiment is related to the issue of generalisability of the RULE approach.

 \subsection{Baseline conditions}
 In all model runs, both the initial Ent population size and initial PP population size were set to 100, and these were distributed randomly across the inner 7x7 m (PPs) and 6x6 m (Ents) of the 10x10 m surface of the simulated environment. PP initial age and mass were chosen at random (to avoid forcing cyclic behaviour in biomass availability): $age\sim U[1,100]$ and $mass\sim U[5,50]$. Ent initial ages were set to zero (so initial training is forced to cover this crucial initial period where risk of death by starvation is high), and their mass to 3.75 kg (the offspring mass of the lightest possible mother). Each Ent is provided (by the mother) with an initial energy store equal to 1 unit (this amount is genetically determined and therefore subject to change when evolution is enabled). Batches of currency $C_4$ (\textit{coins}) are dropped into the environment (materializing at a random height, $\sim U[0.05, 0.1]$, and falling to the surface under simulated gravity) at regular intervals, with a batch of $n_{coins}$ coins per 10 s (referred to as a `coin cycle'). Coins persist on the surface for 20 s (during which time they may be picked up) before disappearing.  Under baseline conditions $n_{coins}=10$.
 The external light source was held constant throughout all simulations at a value of 1. The simulation halted if the Ent population reached zero, and ran no longer than the maximum time limit of 25,000 \textit{s} for all experiments.

\subsection{Preliminary stage}
\label{sec:Preliminary stage}
\paragraph{Baseline policy training and population persistence}
The first step involved choosing (by trial and error) a set of initial values for genes and reward coefficients, and running RL training for a sufficient number of steps to create a robust initial policy under baseline conditions. During the training process, if all Ents in the population died, the environment was reset. Eventually, as the behaviour of the Ents improved under training, reproduction offset the loss (death) of Ents, and the population survived for longer. What counts as a `robust' policy was determined using two attributes:

{(1) Persistence:} the ability to maintain a finite population of Ents over time under benign conditions; this requires sufficient rewards for at least food consumption (for energy supply) and reproduction. The reward coefficients found to meet this requirement were $\theta_{consumption}=0.3$ and $\theta_{reproduction}=1$. In the same spirit, the initial genes chosen for PPs make them palatable to Ents. Relatively small coefficients were assigned to changes in currencies ($\theta_{1,2,3}=0.1$), to encourage currency transfers but to ensure currency dynamics did not initially dominate eating and reproduction during the initial period. The reward for being bitten by another Ent was set to $\theta_{consumption} \cdot \theta_{Pain}=-0.3$ and the reward for dying before the genetically-determined maximum lifespan (from a lack of mass, due either to starvation or predation) was set to $\theta_{Pain}=-1$.  It is unlikely these initial reward coefficients and gene values are optimal in any sense and this was not the goal, but they were sufficient to produce stable populations from which the dynamic updating processes could proceed.

{(2) Coin collecting:} The second necessary attribute was that the Ents should expend effort collecting currency $C_4$ (`coins'). Coin collection is a necessary part of the experiment outlined below and this behaviour can only be achieved if Ents are sufficiently rewarded. The reward for $\Delta C_4$ (collection of a single coin) was set to $\theta_4=0.3$.

\paragraph{Baseline reward expectations}
Following training of the baseline policy (with fixed $\theta$ values) reward expectation distributions, $E_i(t)$, were created, as required for later application of the RULE algorithm. To achieve this, the model was run with evolution disabled (offspring received an exact copy of their mother's genes), reward coefficients ($\theta$) held constant, and Ent behaviour driven by the baseline policy (in inference mode, with no learning). Populations of Ents and Primary Producers were initiated (parameter values equal to those during initial training) and all expectation values were initialised to zero. The model was run for 10,000 s, updating reward expectations for each offspring according to RULE, providing sufficient time for the expected reward distributions $E_i(\tau)$ to stabilize at consistent values (see Appendix A.2.4 for further details). These distributions, together with the trained policy for behaviour, provide the necessary ingredients for the experiments to proceed.

\subsection{Experiment design}
Assessing whether Ents can survive the coins challenge involves controlling combinations of three factors: evolution (of genes), learning (continuously updating the policy through training), and reward updating with RULE (which can only happen during learning) - see Table \ref{experimentsTable}. Note that the learning condition here does not involve resetting of the environment if the population dies. The population is expected to be replenished through reproduction, so if the population collapses that is the end of the experiment. Evolution can be disabled by giving offspring an exact copy of their mother's genes (and disabling mutation); learning can be switched-off by running the Ent behaviour model in inference mode rather than training mode; finally, reward updating (even if learning is enabled) can be switched off by setting $\delta_E=\delta_{\theta}=0$ during all reproductions.

\paragraph{Experiment 1: Baseline coins sensitivity}
The baseline sensitivity of the population to coin rate was measured under otherwise standard conditions, using Ents with no evolution or learning (and no reward updating), primarily to confirm whether a 'fatal dose' of coins exists, and to estimate its value (to inform subsequent experiments and interpretations).The experiment involved 10 repeat runs at each of 10 different (constant) coin rates (from 0-80 coins/10 $s$). Each simulation was run for 10,000 $s$, as this was deemed sufficient time to asses with confidence whether a populations could survive.

\paragraph{Experiment 2: Coins challenge}
Twelve model runs were used in this experiment: 6 combinations of experimental conditions (learning, reward updating and evolution - see Table \ref*{experimentsTable}) with increasing coin rates, plus a control for each condition with coin rate constant at baseline conditions. 
Under the `increasing coin rate' condition, two rates were used: \textit{higher} ($n_{coins}(t)=10+0.014t$) and \textit{lower} ($n_{coins}(t)=10+0.0097t$), where $t$ is simulation time (\textit{s}), meaning the coin rates reach 150 and 107 coins per cycle respectively, by 10,000 s.
\begin{table}[h!]
	\begin{flushleft}
		\begin{tabularx}{0.79\textwidth}{llll}
			\toprule
			Experiment & Evolution & Learning & Reward updating \\
			\midrule
			$\text{[2a]}$ Inference-only & False & False & False \\
			$\text{[2b]}$ Evolution-only & True & False & False \\
			$\text{[2c]}$ Learning-only & False & True & False \\
			$\text{[2d]}$ Evolution + Learning & True & True & False \\
			$\text{[2e]}$ Learning + Reward-updating & False & True & True \\
			$\text{[2f]}$ Learning + Reward-updating + Evolution & True & True & True \\
			\bottomrule
		\end{tabularx}
		\caption{List of experiment conditions. In each case, there are three coin rate conditions, (i) constant (10 coins / cycle), (ii) Increasing from 10 to 107 coins / cycle, (iii) Increasing from 10 to 150 coins / cycle.}
		\label{experimentsTable}
	\end{flushleft}
\end{table}

The RULE algorithm will be deemed successful if the Ent population with RULE enabled can survive increases in coins which are fatal to populations under the same conditions with RULE disabled.
 
\paragraph{Experiment 3: Vitamins and poisons}
The purpose of Experiment 3 is to assess the extent to which explicitly `helpful' or `harmful' behaviours can be amplified/abandoned by reward updating. This is in contrast to the implicit (opportunity) cost associated with picking up otherwise neutral coins in Experiment 2. In Experiment 3, the effects of picking up coins are extended to (i) an instantaneous change in the energy stored by the Ent ($\Delta{C_0}$), and (ii) a multiplier of the probability of reproduction. Under these changes, `coins' are re-conceptualised as`Vitamins' and `Poisons', defined by: 
strong vitamin {$\Delta{C_0}=1$,$P=5$}, 
weak vitamin {$\Delta{C_0}=0.05$,$P=1.25$}, 
strong poison {$\Delta{C_0}=-2$,$P=0.1$}, 
weak poison {$\Delta{C_0}=-0.05$,$P=0.5$}. 
To pass this test, rewards for picking-up vitamins (if truly beneficial) should increase, and rewards to poisons should decrease, from the baseline value of $\theta_4=0.3$.

\paragraph{Experiment 4: Dormant rewards}
The purpose of this experiment is to assess whether (under RULE) Ents can successfully adopt/ignore beneficial/harmful activities of which they have had no previous experience. This is in contrast to Experiment 2 and 3 which test the ability to modify already-learnt behaviours in the face of new environmental conditions. The introduced conditions chosen were the coins and vitamins of Experiment 3. Experiment 4 therefore required starting the baseline training process from the beginning to establish a persistent Ent population without coins having been present, and without coin collection having been learnt. Experiment 4 began with this persistent population, with the coin reward coefficients $\theta_4$) set to a low value of $0.001$, and with expected reward distributions for coin rewards initialised randomly within the range [-0.001, 0.001]. In two separate experiment runs, (1) coins, and (2) vitamins, were introduced at a linearly-increasing rate ($n_{coins}(t)=10+0.014t$, the higher rate used in the coins challenge).

\section{Results}
\begin{figure}[h!]
	\includegraphics[width=6.225in,height=2.14629in, angle=0]{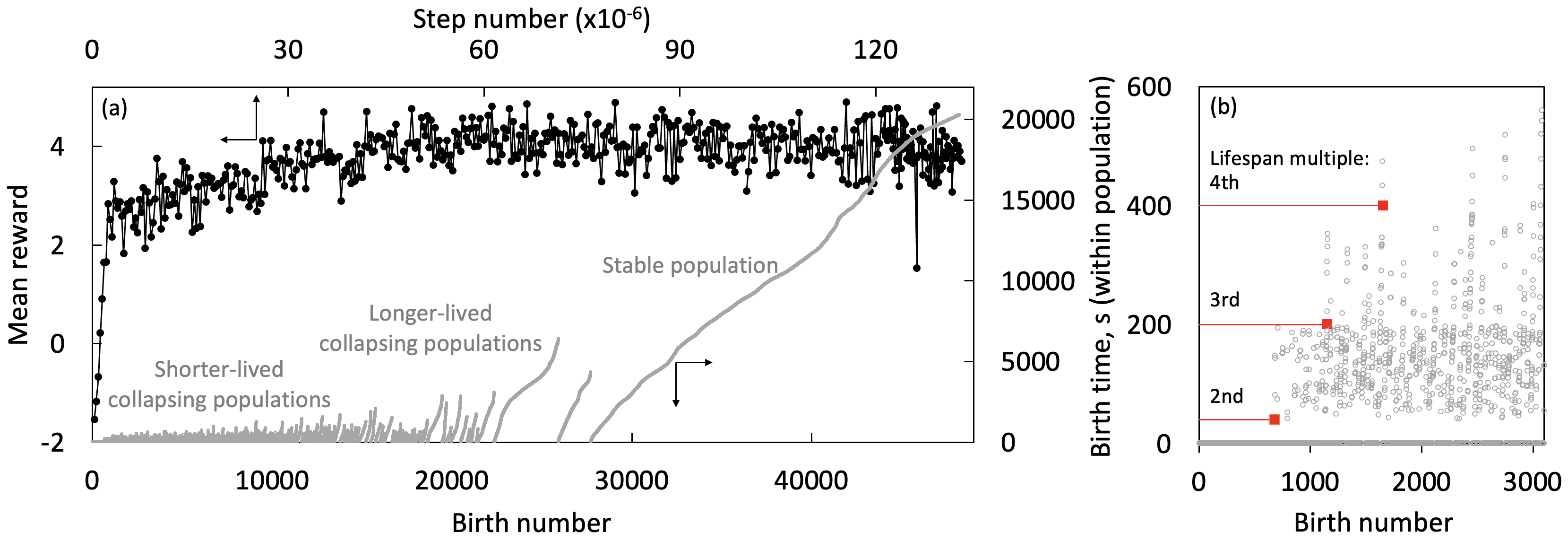}
	\caption{(a) Black symbols indicate the total reward obtained at each step during the initial policy training. Data in light grey are the birth number of each Ent born during the training period, plotted against the time since its population was initiated (which occurs each time the population collapses). Vertical plotting data indicate bursts of reproduction, leading to populations which ultimately collapse; more diagonal data indicate a more consistent birth rate. Part (b) shows the first 3,100 births in greater detail, indicating the points in training where multiples of the maximum lifespan (200 s) are observed, indicating increasingly successful Ent survival and reproduction. Below "2nd", no successful offspring are created; above "2nd" are second generation Ents; "3rd" and "4th" are not strictly generation numbers, but indicate increasing population persistence as Ents are born at times which can only be reached due to multiples of the maximum lifespan.}
	\label{Fig_training}
\end{figure}

\paragraph{Preliminary stage results}
During baseline training (with repeated periods of learning and re-setting the environment after Ents population collapse), Ents successfully learnt a policy which allowed them to survive in their environment, maintaining a persistent population through reproduction. At a minimum this required Ents to successfully identify and then move to targets, eat and reproduce. The time series of rewards obtained during initial training is shown in Figure \ref{Fig_training}, along with data on the succession of Ent populations observed. During the early phase of training (0-30e6 steps), Ent populations collapsed after short periods as individual Ents had not learnt to eat and reproduce. As these and other skills were acquired through training, populations persisted for longer, producing successive generations, (Figure \ref*{Fig_training}(b)) eventually resulting in a stable population (Figure \ref*{Fig_training}(a)) which met both the persistence and coin-collecting criteria outlined above. In subsequent experiments involving continuous learning (a requirement for RULE) the PPO algorithm (with constant learning rate) was found to be stable, with no evidence of  reward/behaviour deterioration, or failure to learn. 

\begin{figure}[h!]
	\includegraphics[width=6.225in,height=3.545in, angle=0]{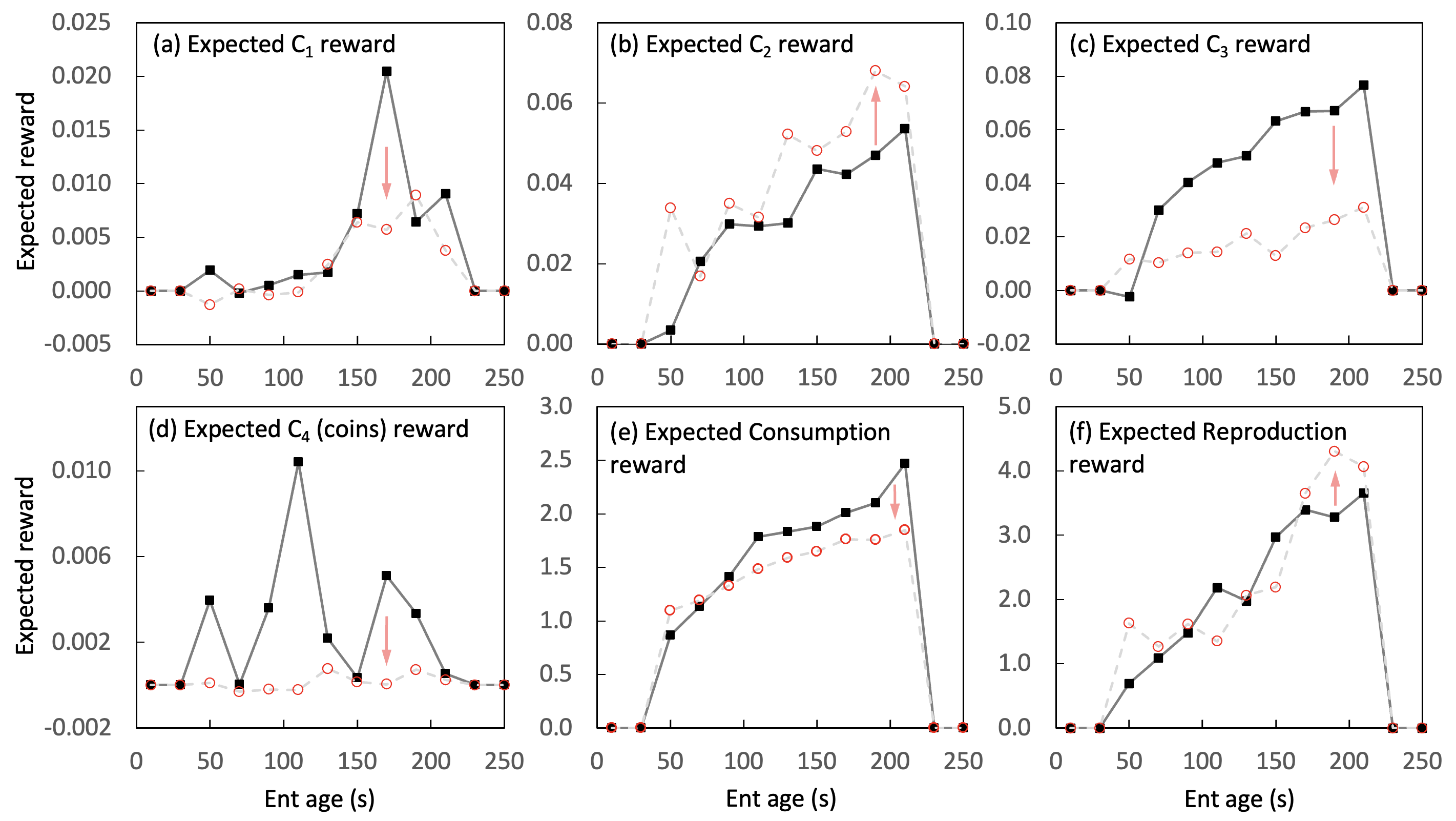}
	%\vskip 0.25cm
	\caption{Expected-rewards distributions $E_i(\tau)$ are shown here for Currency, Consumption and Reproduction reward coefficients.  Black squares indicate the mean values (at each time point) from the last 500 Ents born in the 10,000 s simulation. These values are the `Baseline reward expectations', following the initial policy training based on the chosen reward function coefficients (red circles in the figures relate to the `coins challenge' and are discussed below).}
	\label{Fig_expected_rewards}
\end{figure}

Results for the `Baseline Expected Rewards' distributions $E_i(\tau)$ are shown in Figure \ref{Fig_expected_rewards} for Currency, Consumption and Reproduction rewards. Variation was observed in the distributions of individual Ents, due to differences in life experience accumulated down the maternal line. However, over the course of the 10,000 $s$ model run, mean $E_i(\tau)$ values (averaged over the last $n=500$ Ent births) converged on the values shown (black squares in Figure \ref*{Fig_expected_rewards}) and were used as the starting distributions for each Ent in subsequent experiments (`Baseline reward expectations'). The red circles in these figures are the $E_i(\tau)$ adjustments following the `coins challenge', and are discussed below.

\begin{figure}[h!]
		\begin{flushleft}
		\includegraphics[width=6.2327in,height=2.365in, angle=0]{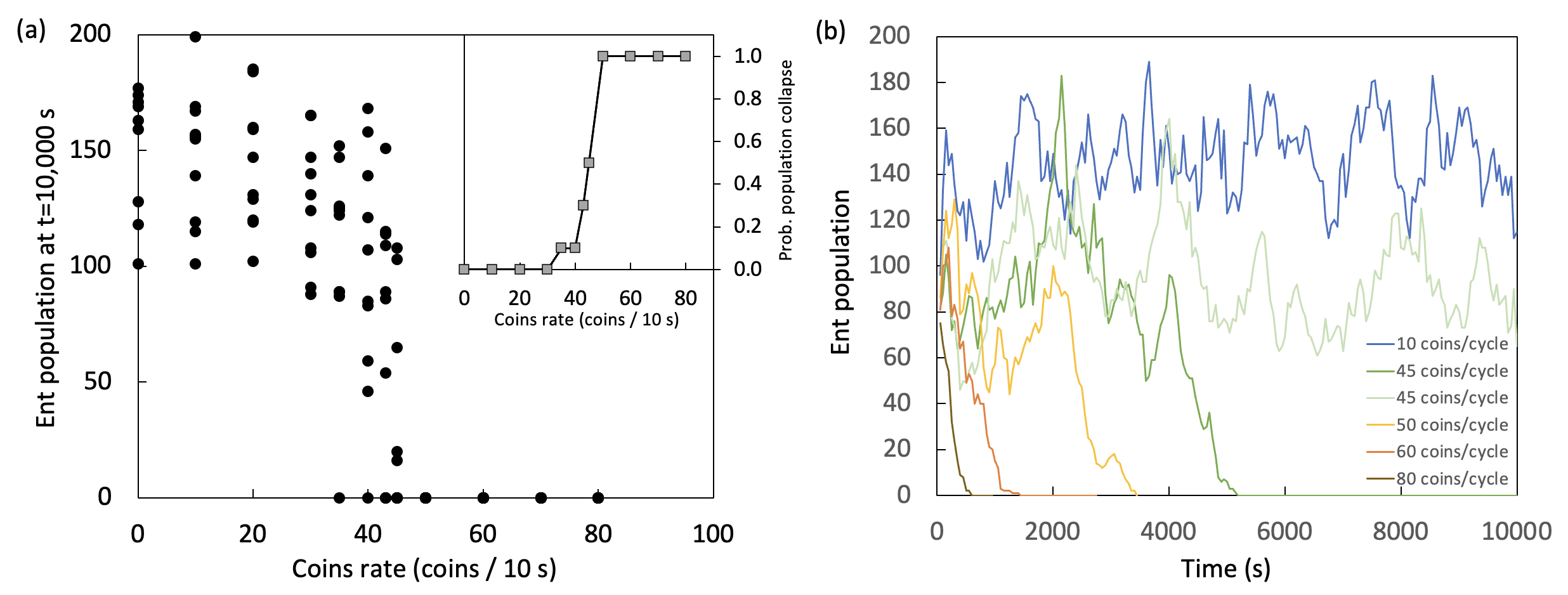}
		\caption{Results from the `Coins sensitivity' experiment. (a) Final population after 10,000 s at the constant coin rate shown in the horizontal axis. The inset shows the probability of population collapse. (b) Example Ent population time series for a range of coin rates.}
		\label{Fig_coins_panel}
		\end{flushleft}
\end{figure}

\paragraph{Experiment 1: Baseline coins sensitivity} Summary results for the preliminary coin-sensitivity experiment are shown in Figure \ref{Fig_coins_panel}. Part (a) plots the final Ent population (after 10,000 s)  against the coin rate. The inset shows the posterior probability of population collapse. Coin rates $\leq30$ do not impact the survival probability of the Ent population; coins rates $\geq50$ always collapsed the population, and those intermediate rates produced increased risk of collapse as coin rate increased (see inset). Between model runs there was broad consistency and example population time series are shown in Figure \ref*{Fig_coins_panel}(b). These results together demonstrate there is indeed a maximum survivable `dose rate' of coins, and that increasing coins over time to beyond this limits would provide a suitable survival challenge.

\begin{figure}[h!]
	\begin{flushleft}
		\includegraphics[width=15.75cm,height=9.32097cm, angle=0]{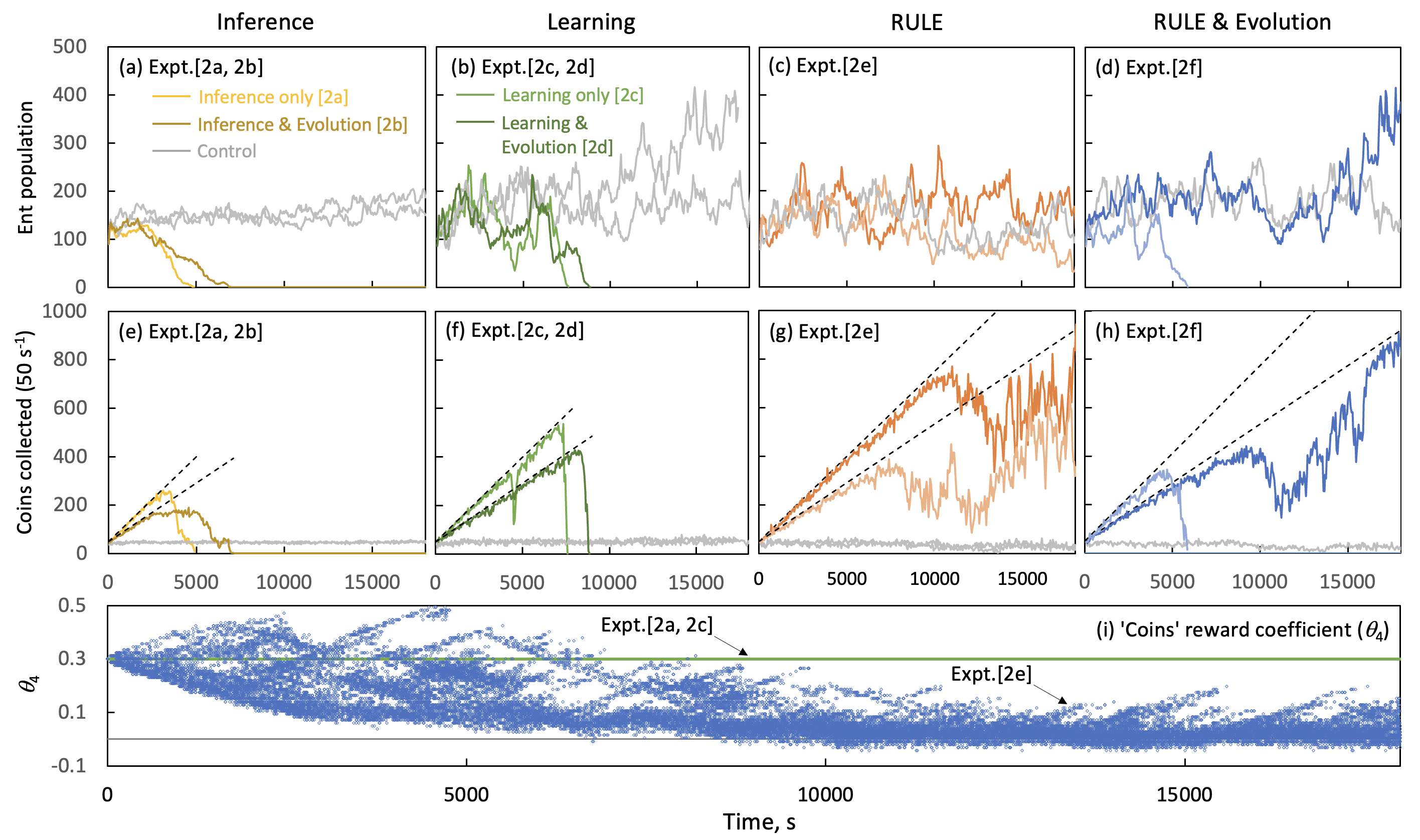}
		\caption{Results from Experiment 2: the `coins challenge'. Grey lines in parts (a)-(h) are data from control conditions in which the coin rate was kept constant at 10 coins/cycle. The dashed black lines indicate the availability of new coins. In part (i) blue symbols indicate coin reward coefficients $\theta_4$ for each Ent offspring, plotted against the time of its birth. Further details are available in the main text.}
		\label{Fig_survival}
	\end{flushleft}
\end{figure}

\paragraph{Experiment 2: Coins challenge} This experiment involves a linear increase in coin-rate (coins dropped at 10 $s$ intervals), under the conditions summarised in Table \ref{experimentsTable}, Experiments[2a-f]. Figure \ref*{Fig_survival} displays representative results from each of these experiments. The top row shows Ent population size over time, the middle row shows the rate of coin collection (across the whole Ent population). In these plots, the dashed lines shows the availability of new coins increasing over time, noting that the actual collection rate may exceed this if Ents drop coins (which survive for 1 $s$ before disappearing) and these are subsequently picked-up. For each of parts (a)-(h), the grey time series is the control condition, in which the coin rate was held constant at 10 coins/cycle. In each of these cases the Ent population survives for the duration of the experiment. The bottom element (\ref*{Fig_survival}(i)) shows how the coins reward coefficient changes over time, plotting the $\theta_4$ value of each Ent offspring, against its birth time, in three representative examples. 

The first column (`Inference', where Ent behaviour choices are made using the fixed baseline policy, with no additional learning) shows the population (a) and the coin collection rate (e) falling to zero as the coin rate increases beyond \textasciitilde40 coins/cycle (\textasciitilde200 coins/50 s), consistent with results shown in Figure \ref{Fig_coins_panel}. Lower coin increase rates are associated with longer-lived populations as it takes longer to reach a 'fatal dose'. For either of the coin increase rates, enabling 'evolution' (Expt.[2b]) makes no significant difference to population survival under the inference-only condition.

Under the `Learning' condition (second column), where Ents actively update their shared behavioural model throughout the experiment, populations survive \textasciitilde50\% longer compared to the `inference' condition, eventually collecting coins at a greater maximum rate. However, a similar overall pattern occurs to the Inference case, in that both population (Figure \ref*{Fig_coins_panel}(b)) and collection rate (Figure \ref*{Fig_coins_panel}(f)) for both coin increase rates, with and without evolution enabled, eventually collapse as coin availability increases. In both cases, coin collection rates grow with coin availability until the point where collection is limited by the dwindling population size.  Under the control conditions, there is no evidence that continuous learning was detrimental to population size, and in some cases there was considerable population growth (as shown in part (b) of Figure \ref{Fig_survival}). The important observation from Experiments [2c] and [2d] is that neither learning (with a fixed reward function) or learning \& evolution leads the population to find a solution to surviving the coins challenge.

The third column shows results for the case of continuous learning with adaptive reward coefficients (RULE). Unlike the previous experiments described, the population levels in these experiments, while variable, remain at broadly consistent levels as the coin delivery rate increases. The coin collection rates initially increase with coin availability under both availability increase rates, but then reduce considerably, with not all of the available coins being collected. This reduction is not caused by population collapse (as in the Inference and Learning conditions) but by changes in Ent behaviour driven by changes in the reward coefficient ($\theta_4$) and in the expected reward distributions ($E(\tau)$, as shown in Figure \ref{Fig_expected_rewards}). The Ent population survives increases in coin rate to levels beyond 250 coins/cycle, meaning reward updating via RULE appears to provide a mechanism for the population to survive the coins challenge. 

The final column (`RULE \& Evolution') where evolution of the Ent `gene' is enabled together with continuous learning and reward coefficient updating, the results depend on the coin increase rate. At the higher of the two coin availability rates, the population collapses relatively quickly as coin availability increases. At the lower rate, the response in terms of coin collection is similar to the `RULE' condition over the first \textasciitilde10,000 $s$, but the population increases dramatically from \textasciitilde12,000 $s$, and the coin collection rate eventually matches the coin availability rate. 

Part (i) of Figure \ref*{Fig_survival} shows the reward coefficient for coins ($\theta_4$) over time, for Experiments [2a,2c,2e]. In the case of Experiments [2a,2b] (Inference) and Experiments [2c,2d] (Learning), all reward coefficients are held constant. In Experiments [2e,2f] (RULE), where reward coefficients are updated actively over time, significant changes are observed (results for both experiments are similar and for clarity, only results for 2f are shown). Each point plotted represents the $\theta_4$ value of an Ent offspring, at the time of its birth. A dominant features is for $\theta_4$ to reduce towards zero, and this is evident from the beginning of the time series. The range of $\theta_4$ values initially increases, going above and below the initial value of 0.3, before becoming more restricted at later times. 
During the `coins challenge' (Experiment 2),  the increasing availability of coins means some Ents collect more, and subsequently give birth to Ents which similarly pick up relatively large numbers of coins. This drives up $\theta_4$ for those Ents and these `lineages' can be seen in the data plotted in Figure \ref*{Fig_survival}(i). At the same time, other Ents pick up fewer than expected coins, and give birth to Ents who have similar experiences. Hence in the early stages of Experiment 2e, there is a diversity of branching $\theta_4$ responses. As coin availability increases further, those Ents who happen to collect more coins are on average less successful at producing offspring, and so the `signal' to the population encouraging greater coin collection is relatively weak. Conversely, those collecting fewer coins will on average produce more offspring with $\theta$ and $E(\tau)$ values which encourage less coin collection. As these effect continue, $\theta_4$ drops significantly over the course of the experiment, to values below those for both $\theta_{consumption}$ and $\theta_{reproduction}$. The outcome in terms of behaviour is that Ents pick up coins if they are in their vicinity, but won’t leave food or reproduction opportunities to get coins. So while coin collecting behaviour continues, many coins are left un-collected. 

\begin{figure}[h!]
	\begin{flushleft}
		\includegraphics[width=16cm,height=13.76cm, angle=0]{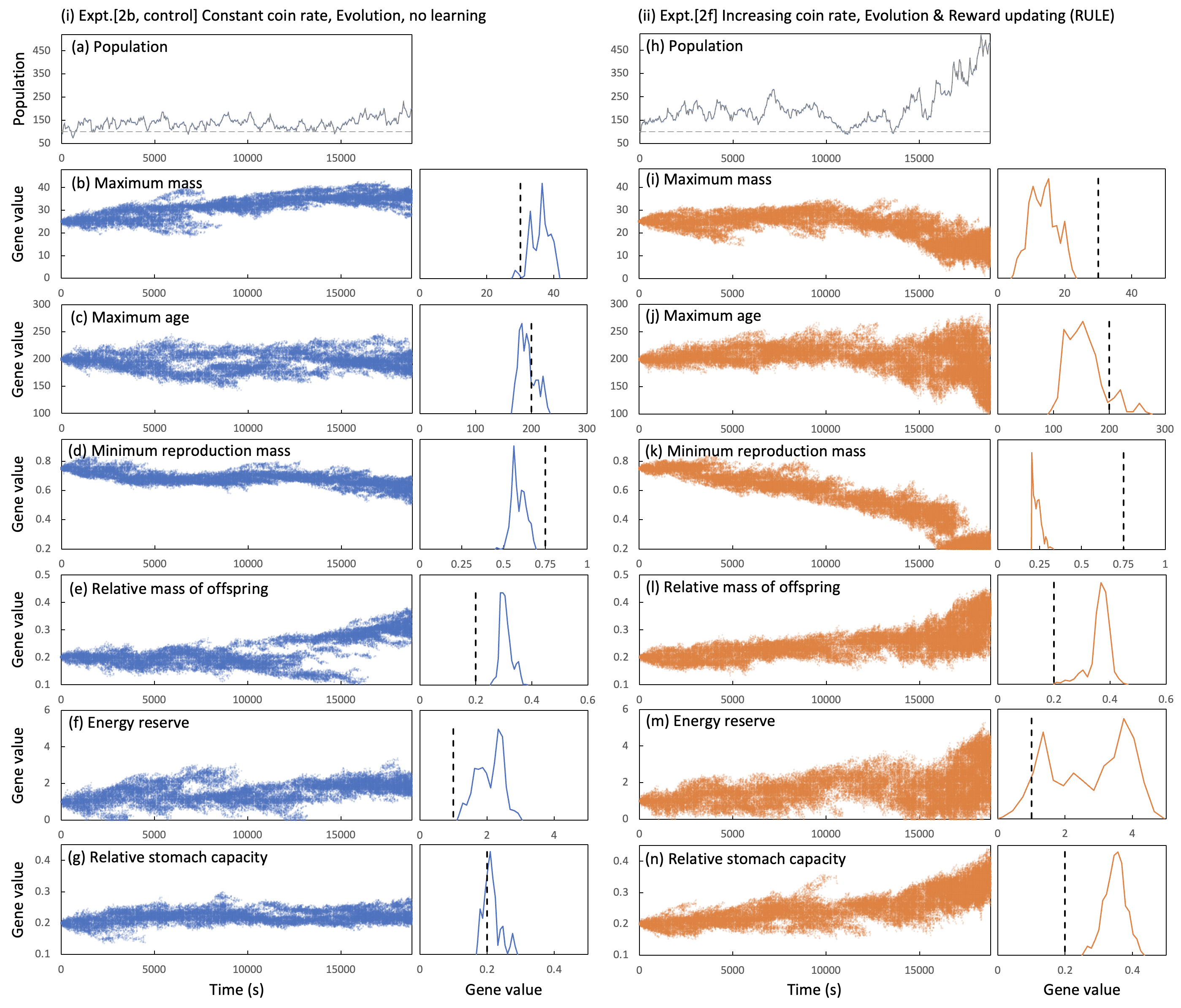}
		\caption{Evolution of gene values. Each point in the time series represents the gene value of an Ent, plotted at its time of birth. The gene value distributions are for the last 500 Ent  births of the experiment, with the vertical dashed line indicating the initial gene value in each case.}
		\label{Fig_evolution}
	\end{flushleft}
\end{figure}

\paragraph{Effects of evolution} 
For experiment [2b], 10 repeats of the control condition were run (evolution was enabled, but no learning or reward updating). Across the results from these runs some gene values showed no consistent changes and drifted to both higher and lower values. In contrast, other gene values showed consistent patterns across all 10 repeat runs. The left-side (blue) panels of Figure \ref{Fig_evolution} show representative results of this latter group. Each data point represents the gene value for an Ent offspring, plotted at its time of birth. The distributions shown to the right of the gene time series are frequency distributions of the gene values from the last 500 offspring of the time series; the dashed line indicates the initial genes value. Amongst these data there is a tendency over time towards the following: larger Ents [evolving from an initial value of 25 \textrightarrow mean value of 36] (b) who keep larger energy reserves [1.0 \textrightarrow 2.1] (f), with consistent (<5\% change) relative stomach sizes [2.0 \textrightarrow 2.1] (g) and maximum lifespans [200 \textrightarrow 193] (c); able to reproduce when smaller (lower body mass) [0.75 \textrightarrow 0.60] (d), giving birth to relatively larger offspring [0.20 \textrightarrow 0.31] (e). These genetic changes yield a stable persistent population.
For comparison, equivalent (orange) plots are shown on the right side of Figure \ref*{Fig_evolution}, generated using data from Experiment [2f] (increasing coin rate, learning and reward updating via RULE \& Evolution). Under these conditions, the evolutionary response has some key differences to the control condition described above: Ents become \textit{smaller}, with lower maximum mass [25.0 \textrightarrow 13.6] (i); their maximum age reduces to a greater extent [200 \textrightarrow 118] (j); minimum reproduction mass is reduced by a much greater amount, with the distributioni running in to its enforced minimum value at 0.2 [0.75 \textrightarrow 0.23] (k) while the relative mass of offspring increases to a greater extent [0.20 \textrightarrow 0.36] (l); relative stomach masses increase dramatically [0.20 \textrightarrow 0.35] (n) and energy reserves (while more variable)  extend to higher values [1.0 \textrightarrow 2.7] (m). 

Proposed explanations for the changes observed during the control conditions and the coin challenge conditions are presented in the Discussion section below.

\begin{figure}[h!]
	\begin{flushleft}
		\includegraphics[width=16cm,height=10.223cm, angle=0]{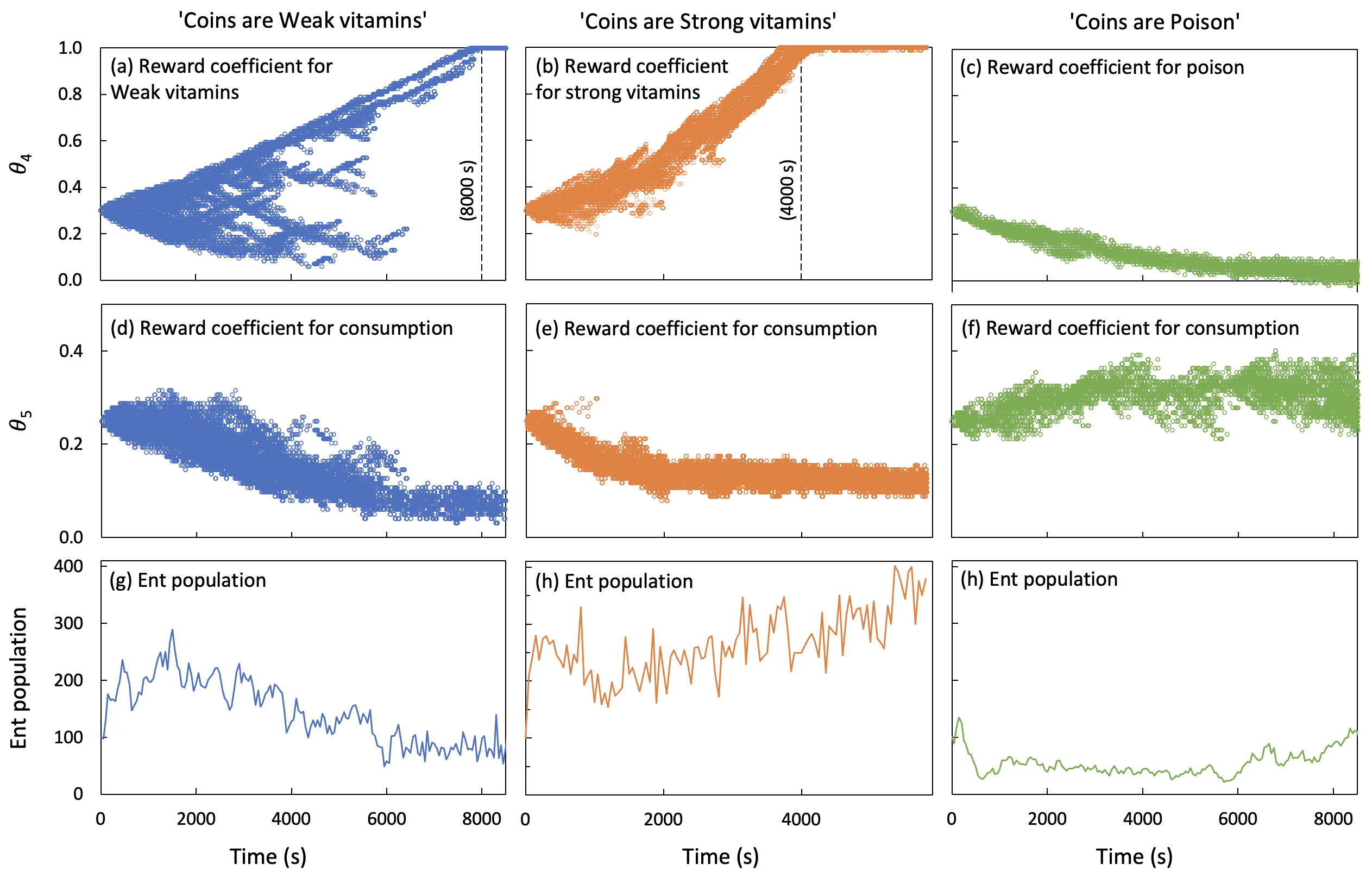}
		%\vskip 0.25cm
		\caption{Vitamins and Poison. Results from three experiments in which `coins' were recast as `vitamins' or `poison'. Picking-up vitamins provides an instantaneous increase in stored energy and an increased probability of becoming reproductive. Conversely, picking up`poison' reduces both stored energy and reproduction probability. The results shown are for cases where RULE updating of reward coefficients is enabled. The positive/negative benefits/costs of vitamins/poisons are reflected in the increased/decreased reward coefficients for `coins' ($\theta_4$).}
		\label{Fig_vitamins_and_poison}
	\end{flushleft}
\end{figure}

\paragraph{Experiment 3: Vitamins and poisons}
Figure \ref{Fig_vitamins_and_poison} present results from Experiment 3, a `coins challenge' experiment (with RULE and no evolution) in which the act of picking up coins was combined with either energy-and-reproductive benefits (coins as `vitamins') or costs (coins as `poisons'). The top and centre rows provide time series of reward coefficients for coins ($\theta_4$) and consumption ($\theta_{consumption}$) respectively; the bottom row shows Ent population time series. Three experimental conditions are shown in the columns: weak vitamins, strong vitamins, poison (see experiment description section above for details). In both `vitamins' experiments, $\theta_4$ increases dramatically while the consumption (eating) reward $\theta_{consumption}$ drops significantly. In both cases reward coefficients $\theta_4$ reach the maximum value of 1, and this happens in \textasciitilde8,000 $s$ for weak vitamins, versus a much faster \textasciitilde4,000 $s$ for strong vitamins. These relative rates are also seen in the reduction of $\theta_{consumption}$ values. Conversely, $\theta_4$ reduces to near zero in the `coins as poison' condition, and this is accompanied by an increase in $\theta_{consumption}$. In all cases, Ent populations remain viable throughout the experiment.  The core relevant observation from these results is that reward coefficients \textit{increase} when coins are (at least superficially) beneficial (vitamins), and \textit{decrease} when coins are harmful (poisons).

\begin{figure}[h!]
	\begin{flushleft}
		\includegraphics[width=14cm,height=14.60cm, angle=0]{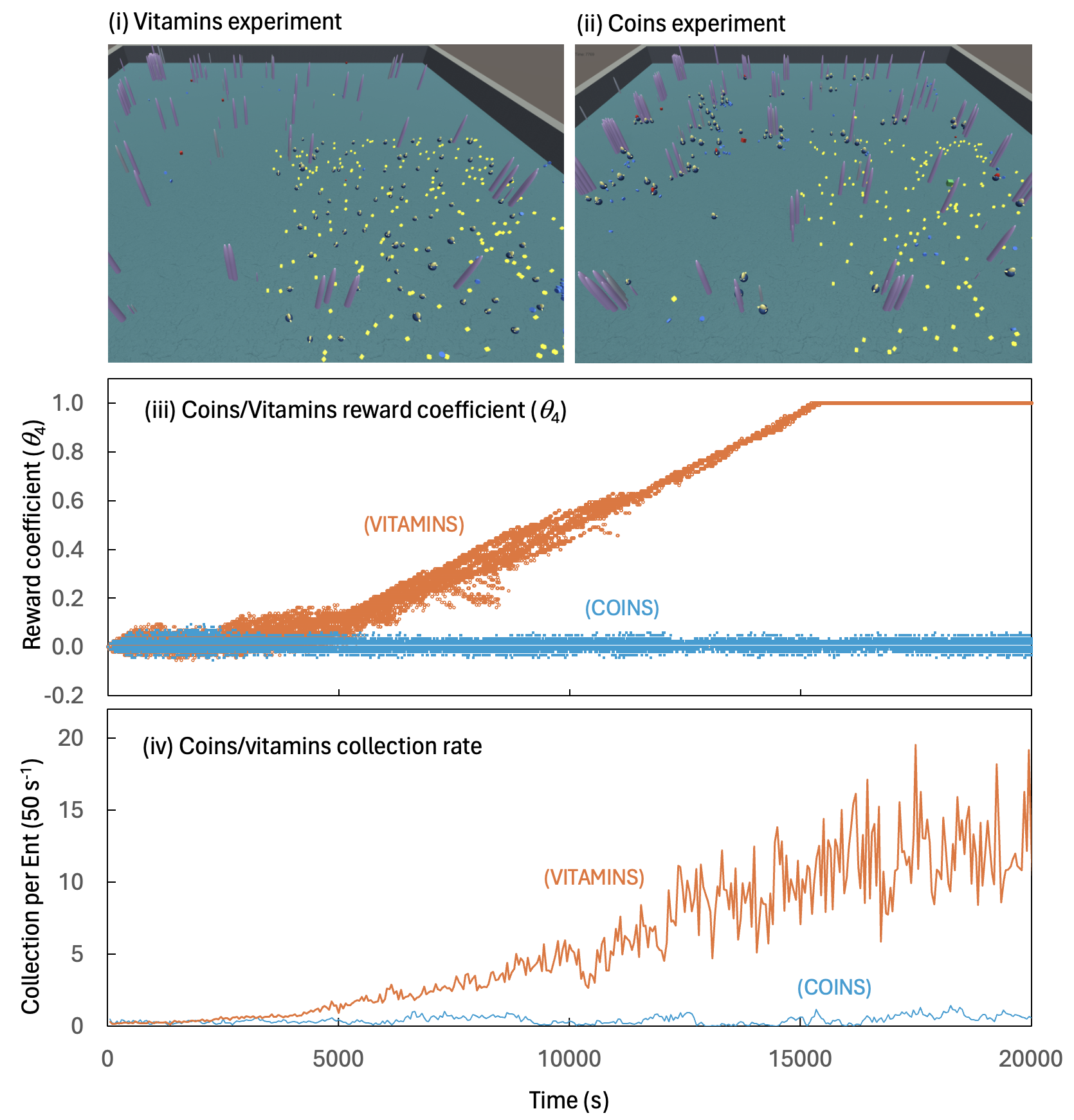}
		%\vskip 0.25cm
		\caption{Dormant reward response to Coins and Vitamins. (i) and (ii) show example snapshots of the distribution of Ents in the environment during each experiment. The clustering of Ents close to Vitamins and the avoidance of Coins is observed. In (iii), each point represents the $\theta_4$ value of an Ent born at the time indicated. (iv) shows the rate (per Ent) at which coins/vitamins were collected over the duration of the experiment.}
		\label{Fig_dormant_rewards}
	\end{flushleft}
\end{figure}

\paragraph{Experiment 4: Dormant reward}
Figure \ref{Fig_dormant_rewards} shows results from Experiment 4 (in which coins are absent from the environment during baseline training, and are then introduced at a linearly-increasing rate during the experiment; RULE is enabled and there is no evolution). Figure  \ref*{Fig_dormant_rewards} parts (iii) and (iv) together show that $\theta_4$ and the rate of \textit{coin} collection remain consistently around zero for the 20 ks of the experiment; conversely, $\theta_4$ values and the rate of \textit{vitamin} collection per Ent show significant increases over time after \textasciitilde2 ks. The upper limit of $\theta_4=1$ is reached at \textasciitilde15 ks and is associated with a stabilised collection rate. The difference in behavioural response to introduced coins and vitamins is indicated in Figure\ref*{Fig_dormant_rewards}: by the end of the experiment, Ents are positioning themselves roughly evenly within the area where vitamins are dropped (ii), and collecting them quickly, while avoiding the area where coins are dropped (i), and showing no consistent interest in collecting them.

\paragraph{Currency and Signals transfers}
Currency and signal transfers are not the core focus of the current experiments, and only brief mention is given here. Transfers of currencies between Ents are observed in the results, and signals were also passed. However, there were no major consistent trends in the data across the various experiments, suggesting these behaviours were not instrumental in surviving the coins challenge, or in surviving the otherwise benign conditions of the baseline experiments, and for the sake of brevity are not reported further.

\newpage
\section{Discussion} \label{Discussion}

\paragraph{Death by coins}
Collecting coins is intentionally frivolous as it wastes time and energy that could be spent on more productive things (eating and reproduction, for example). Baseline coin rewards were set relatively high at the outset, to promote coin collection, with a small and fixed supply rate of coins promoting competition amongst the Ents (as was observed). With fixed rewards, high rates of coin supply are fatal to the Ent population: greater coin availability per Ent means more time collecting them, hence the sometimes sharp populations collapse as more coins are collected by ever-decreasing numbers of Ents. Results show no evidence that (continuous) learning, with fixed reward coefficients, can solve this problem, as learning is always in service to the same reward function. Behavioural policies that would have resulted in higher rewards (with the same reward function), e.g. collecting \textit{fewer} coins in order to preserve life (hence avoiding the -1 reward for early death) plus eating and reproducing more, were not discovered during training before the population collapsed. Given large numbers of repeated attempts, such policies may have been discovered, but to do this in linear time appears beyond current learning approaches. Indeed, if this were possible a single high-level sparse reward may indeed meet the challenges posed in this work (although this of course does not avoid the issue of choosing what that high level reward should be).

\paragraph{Testing RULE: the coins challenge}
The RULE algorithm is proposed as an approach to address the problem of fixed reward coefficients: parents (mothers) modify the values of reward coefficients and expected rewards and expected reward distributions passed on to their offspring, based on their own experiences. During continuous learning across the population, the policy is updated to maximize rewards and the modifications parents make to  $\theta$ act to encourage their offspring to behave more like the parents. A diversity of experiences across the Ent population, and hence differences in the `success' of individuals, is exploited to promote behaviour which is good for the population. The process is a dynamic majority vote to reinforce beneficial modifications to reward coefficient values, but also to leave open the possibility of reinforcing different behaviours. For example, in results for the second half of Experiment 2, the average $\theta_4$ value is close to zero, but small sub-populations continue to have different coin collection experiences and produce short-lived `exploratory' lineages, as excursions from the main trend. These excursions die out if the associated experience is not beneficial but it is this exploration of rewards that is needed to find new behavioural solutions to novel conditions (as seen in Experiments 3 \& 4.).
The evidence supports the conclusion that dynamic beneficial reward updating is possisble. This is true for coins, even though in themselves they are not harmful, and their impact on Ents is indirect, via opportunity costs. RULE should work equally well, or even more `purposefully' when faced with more direct impacts. To this end, Experiment 3 (`vitamins and poisons') was focused on understanding how the nature of the response to collectable objects depends on their properties. 

\paragraph{Testing RULE: vitamins and poisons}
Vitamins provide clear benefits to Ents (increased energy and reproduction rate), their collection makes sense, and the results show as vitamins are made available, RULE increases $\theta_4$; further, the associated increase in collection rate which follows reflects the strength of their benefits. The response to stronger vitamins is faster and more coherent, and there is little `exploration' of lower $\theta_4$ values (Figure \ref{Fig_vitamins_and_poison}(b)). For weaker vitamins, where the benefits are more marginal, there is more `exploration' of lower $\theta_4$ values before higher rewards eventually win out (Figure \ref*{Fig_vitamins_and_poison}(a)). A greater level of `deliberation' in less straightforward cases is perhaps inevitable given the presence of conflicting reward-updating signals. In both cases $\theta_4$ values reach their maximum limit of 1 and there is a major change in Ent behaviour. The Ents consumption of PPs and processed material drops significantly, as they become reliant on vitamins (reward coefficient $\theta_{consumption}$ values drop significantly in both cases). This behavioural response to vitamins represents a major departure from the baseline behaviour of the Ents.
The opposite response is seen in Figure \ref*{Fig_vitamins_and_poison}(c,f): reduced $\theta_4$  and increased $\theta_{consumption}$. Poisons have a negative impact on Ent energy levels and reduce reproduction rates. Under RULE, they are almost entirely abandoned (very few poisons are picked up) and Ents navigate between/away-from poisons if they come across them.
As seen in the results for the `coins challenge', the diversity of experiences across the Ent population is exploited by RULE as it responds to both beneficial (collecting vitamins) and harmful (collecting poisons) behaviours, by appropriately modifying rewards and therefore learnt behavioural responses.

\paragraph{Testing RULE: dormant rewards}
The reward function components used in the present simulation, while defensible in that they produce persistent populations, are somewhat arbitrary, and this inevitably limits generalisability: only the reward components present can be updated and these may not necessarily apply to future changes in the environment. Two options to tackle this problem are discussed: (1) Predefined `dormant' rewards, (2) Dynamic reward function expansion. 

Option (1) involves increasing the fixed number of components in the reward function, with a subset of these being redundant at first (in the sense they contribute nothing to the total reward as the phenomenon they reward is absent and the coefficient is set to zero). In this case, at some point in the future RULE might be provoked by the environment in to `switching-on' the rewards (increasing the absolute value of the coefficient), thus having it contribute more or less to the ongoing learnt behaviour.
Option (2) calls for a mechanism to recognise new `rewardable' phenomena and to dynamically add relevant components to the reward function (for example, when phenomena are above some threshold level of surprise). Option (2) approaches Option (1) as knowledge of possible future rewards increases. The question of how to implement option (2) remains open. 

Experiment 4 provides a test of how well RULE responds in the presence of a dormant reward (Option (1)), in this case for coins and vitamins. When a persistent Ent population experienced coins for the first time (with $\theta_4\approx0$), a small number of coins were picked-up. As coin availability was increased (linearly) over time, coins continued to be picked-up only at a low rate and rewards did not change significantly: coin collection was not `switched-on' to any significant degree. Coins provide no benefit and these findings are consistent with those of Experiment 2. Conversely, after vitamins were first experienced, the dormant reward showed signs of being `switched-on' after  \textasciitilde1 ks, with rewards and collection rates increasing significantly ($\theta_4$ eventually reaching its maximum value). These observations suggest RULE has the potential to respond appropriately to novel conditions, setting reward coefficients to relevant values which reflect the benefit of the action.  These results underscore how `artificially' high the baseline value for $\theta_4$ was set, and that given the chance to either start at $\theta_4=0$ and discover coins for themselves, or to start at $\theta_4=0.3$ and adjust to the reward, Ents using RULE settle on $\theta_4\approx0$.

\paragraph{Impacts of Evolution}
With broadly stable conditions, no evolution of the primary producers, and only a single species of Ent, there is limited evolutionary pressure in the present set of experiments. Nonetheless, if the initially chosen baseline gene values were sub-optimal (which is likely), evolution would be expected to fine-tune these values over time, and this is essentially what is observed. Each of the consistent gene-value changes observed under baseline conditions can be interpreted as aiding survival (larger Ents with greater energy reserves who start to produce larger offspring earlier in their lives). 
Under the coins challenge experiment, evolutionary changes alone were not sufficient for the population to survive, but in combination with learning and reward updating (under the slower of the two coin increase rates), systematic changes in gene values did produce Ents with characteristics well-suited to surviving high-coin-rate conditions, and resulted in the largest populations recorded across all model experiments. However, as shown in Figure \ref*{Fig_survival}(d), under the same conditions but with the higher rate of increase in coin availability, the population collapsed (results which were consistent across multiple repeat runs). In ecological settings, populations may experience a lag in adaptation when conditions change rapidly, meaning relatively slowly-changing traits which were once beneficial become harmful under new conditions. In the present case, population collapse happens before large systematic changes in mean gene values have happened, but there is a significant increase in the \textit{spread} of gene values (\textasciitilde25\% mean increase in relative standard deviation). There is a potential asymmetry present in that random combinations of gene values which aid survival have relatively small effects (as seen in Experiment 2a,b), whereas those which hinder survival have enough of an effect, in the presence of high coin-stress, to reduce the population. Once the population starts to decline, positive feedback (increasingly higher availability of coins per Ent) takes over (as discussed above). It appears then that if RULE can provide resilience to coins, Evolution fine-tunes the Ents, but if the increase in environmental stress is too fast, the negative aspects of gene value variation collapse the population before RULE can save it.

\paragraph{Reward hacking}
The goal of RL is to tune a policy to maximizes long-term rewards, and during continuous learning the RULE algorithm modifies reward coefficients based on policy-driven behaviour. In principle then, the training process could hijack the adjustment of actions in order to drive a modification of rewards, thus boosting the rewards obtained (reward hacking).  In fact, multiple simultaneous interacting optimisation processes are at work in some experiments: implicit evolutionary optimisation of gene values, explicit maximizing of rewards through policy updating, and the RULE-based reward coefficient updating. Potential exists for any of these to amplify or dampen each others' impacts.
Constant baseline conditions, with RULE and evolution enabled, and all but one (reproduction) of the reward coefficients set below their maximum value of 1.0, provides a good base from which to test for reward hacking. In the results from this case (the control condition of Experiment 2f), no systematic increase in reward coefficients is observed. It is possible that survival needs impose strong constraints on what behaviours can be rewarded (increasing rewards for one activity potentially means less of another), and that this keeps reward hacking in check. There are however other observations which paint a more nuanced picture. During the coin challenge (Expt. 2e; RULE with no evolution), $\theta_4$ is reduced to a level where coin collection still happens, but at a much-reduced level. That is, the reward from coins is not lost, but it is reduced to a point which does not threaten the population (in stark contrast to the case of strong poison, where collection was halted altogether). The response to Experiment 2e created a population largely un-interested in coins, living outside the area where coins were dropped, with a small fraction of the Ents collecting coins at relatively low rates in the coins area. When evolution is added to this condition (Expt. 2f), the body characteristics of Ents are changed significantly, resulting in a very large population of shorter-lived Ents who as a population are able (by the end of the experiment period) to collect very high levels of coins without collapsing the population. This `solution' to the coins challenge is therefore successful in both producing a surviving population and also maximising the reward from coins.

\paragraph{Wider application of RULE}
To what extent can RULE be used in other multi-agent contexts?   
RULE operates at the population level by exploiting variation in individual experiences, and works essentially because successful/unsuccessful individuals reproduce/die.
The current context approximates a simple ecosystem, where Ents face a constant survival challenge and one potential solution to generalising RULE is to cast other situations as analogues of survival challenges.
For example, economic agents may have money in place of energy, costs such as wages as base metabolism, acquisitions as foraged food, sales as consumption, expansion (not including evolution) as reproduction, bankruptcy as death, etc.. Such analogies can become rather strained, but with populations of competing entities which can in some sense reproduce and die, application fo RULE may be fruitful. 
Where individual agents in the simulated system are not of a type which produce offspring (or have enforced maximum lifespans), a different mechanism is required. Agents could, for example, periodically replace themselves with updated versions (effectively reproducing `in place') when some condition was met (e.g. energy or other resource).
In this case, comparison of accumulated rewards to expected rewards over a rolling time window might be used to calculate adjustments to $\theta_i$ and $E(\tau)$.
If agents `die' in such a scenario, new agents would need to be added according to rules which made sense within the context of the specific system.

\paragraph{Hidden goals}
In the introductory sections it is argued that the goals of autonomous real-world agents are typically hidden, and that survival challenges guide rewards to satisfy those hidden goals. Since RULE can adjust reward coefficients, it is natural to consider whether the resultant reward function might be interpretable in terms of the hidden goals it serves. 
Updated rewards elicit individual behaviours which result in emergent consequences at the population level (as in the present case, agents operate within the context of other agents). 
To take an extreme hypothetical case, imagine Ents using RULE see an increase in the reward for consumption, which rises to the maximum value of 1. At the same time, all other rewards fall to low levels. 
Interpretation of the goal may appear simple: maximizing consumption of food is the most important factor in survival. 
However, it may be that when all of the population focusses on eating, there is a secondary effect such as out-competing a second species which would otherwise have competed in some other way (e.g. competition for space). The imperative driving the true goal therefore is reducing competition for space. This was achieved through competing for food, but could feasibly have been achieved in some other way. Eventhough this is relatively simple, the true high level goal is not easily derived from the reward function, or at least not uniquely. 
When there are multiple reward components and more complex behaviours, it seems likely identifying goals would become more difficult. It is argued here then that hidden goals remain hidden, even when the results of reward function updating are known.

\section{Conclusion}

Central to this work is the idea that the goals and reward functions of real-world entities, which we may wish to represent in models, are hidden from the observer (and most likely also from the entities themselves).
These reward functions that guide behaviour are in some sense chosen, or emerge, in real-world systems and adapt over time as the environment changes.
If out of a necessity to elicit desired aspects of the entity's observed behaviour, we decide on a fixed reward function, we risk pre-defining incorrect behaviour in novel situations, or cases for which we have no observational data. Fixed reward functions also limit the adaptability of entities to novel situations.
Hence there is a need to develop means for modelled entities to endogenously shape their own reward functions over time, particularly in response to changes in their environments, and thereby update their own behaviour for their own benefit. 
If our aims is to reproduce the dynamics of real-world systems, entities must also re-shape their behaviour in \textit{linear time}, without the chance to learn by resetting their environments when they fail.  
The evidence presented from simulations of entities (Ents) within a simplified ecosystem-like environment shows that continuous Reinforcement Learning, together with continuous updating of the reward function using the RULE algorithm (Reward Updating through Learning and Expectation), can achieve these goals.
The algorithm exploits the diversity of experience across the Ent population to endogenously reinforce beneficial behaviours and reduce those which are harmful. 
This was demonstrated first by Ents `choosing' to abandon previously learnt and positively-rewarded behaviours when they became harmful.
The population also responded appropriately to previously unseen objects, by creating positive rewards for (and collecting) beneficial `vitamins', and holding the reward at zero (and not collecting) `coins' with no intrinsic benefits when these were introduced in to their environment.
The ecosystem-like context of the present experiments was chosen as a `toy' example because it is conceptually simple and contains generic properties found in many systems: multiple competing agents, scare resources, a spatial domain, costs, exchange of information, etc.. Applying RULE more generally, in which multi-agent problems are cast as survival challenges, is likely possible and a potentially useful focus for future work.

	\newpage
	\bibliographystyle{unsrt}  
	\bibliography{references}  
	
	\newpage
	\appendix
	
	\renewcommand{\theequation}{A.\arabic{equation}}
	\setcounter{equation}{0} % Reset the equation counter to 0 in the appendix
	\section*{Appendix}
	%\addcontentsline{toc}{section}{Appendices} % to add to table of contents

	\section{Additional model description}
		Broad descriptions of the simulated environment, its model components and their behaviours are provided in the main text. Finer details of aspects not covered there are provided below.
		
		\subsection{Primary Producers}\label{sec:PPGene}
		Each Primary Producer (PP) updates its state every $P= 25\Delta t = 0.5 s$ (the model update frequency is $50 s^{-1}$; $\Delta t=0.02 s$). The characteristics of each PP are determined by its gene values. In the present simulations, evolution is not enabled and when reproducing each PP simply passes on an exact copy of its genes.
		
		\textbf{Genes}
		Each PP has its own set of genes which determine its characteristics, as outlined below
		
		\begin{table}[h!]
			\begin{tabular}{lllll}
				\toprule
				Description & Symbol & Units & Range & Initial value \\
				\midrule
				G[0] maximum biomass & $m_{ppMax}$ & kg & $>0$ & 50\\
				G[1] `Lock' L1 [colour, r] & $L_1$ & - & $[0, 1]$ & 0.9\\
				G[2] `Lock' L2 [colour, g] & $L_2$ & - & $[0, 1]$ & 0.7\\
				G[3] `Lock' L3 [colour, b] & $L_3$ & - & $[0, 1]$ & 0.9\\
				G[4] Probability increment per 1\% of accumulated mass & $r_{inc}$ & - & $[0, 1]$ & $5 \cdot 10^{-5} $\\
				G[5] Initial reproduction probability & $r_0$ & - & $[0, 1]$ & 0\\
				G[6] Maximum age & $a_{max}$ & s & $>0$ & 100\\
				G[7] Gene copying noise & $z_{pp}$ & - & $[0, 1]$ & 0\\
				G[8] Energy harvesting coefficient & $c_{s,m}$ & - & $>0$ & 2\\
				G[9] Body mass energy density & $\rho_{E_{pp}}$ & E/kg & $>0$ & 5\\
				G[10] Maximum reproduction probability (per dt) & $r_{pp_{max}}$ & - & $[0, 1]$ & 0.005\\
				G[11] Exclusion radius to height ratio & $h_{ex}$ & - & $>0$ & 3\\
				G[12] Growth rate (kg growth per kg of body mass per s, when s=1) & $g_{pp}$ & $s^{-1}$ & $>0$ & 0.5\\
				G[13] Pre-growth pause & $t_{pause}$ & s & $>0$ & 5\\
				G[14] Inner exclusion ratio & $r_{in}$ & s & $>0$ & 0.2\\
				G[15] Proportion of seeds on surface & $p_{surf}$ & - & $[0,1]$ & 0.05\\
				\bottomrule
			\end{tabular}
			\caption{Gene values for Primary Producers . Gene index numbers are shown. Genes [1,2,3] are jointly the \{r,g,b\} colour tuple (expressed as ratios) for the PP outward appearance and the `lock' values L1-3 which determine palatability. As PP evolution is not enabled in the present case, the initial values shown remain fixed throughout all model runs.}
			\label{tab:table4}
		\end{table}
		
		\textbf{Reproduction via seed production}  Each PP produces a fruit within any given update step with reproduction probability $r$, which is initialised at $r_{pp_{0}}$ and increases over time if \(r_{pp} < r_{pp_{max}}\) where \(r_{pp_{max}}\) is a genetically-set maximum reproduction probability.
		The increment increases to a maximum, as 
		\begin{equation}
			{\Delta}r_p = \frac{m_{pp}}{m_{ppMax}} \cdot r_{inc} \cdot p + n_{bites} \cdot r_{inc}
		\end{equation}
		where, $m_{pp}$is current PP mass (kg), $m_{ppMax}=50 kg$ is the genetically-set maximum mass, $ r_{inc}=5 \cdot 10^{-5} dt^{-1}$ is the reproduction probability increment, $n_{bites}$ counts the number of times the PP has been bitten by Ents since the last fruit production. This provides feedback from Ent behaviour to PP behaviour, relevant to co-evolution of Ents and PPs, but has limited impact at present as evolution of PP gene values is not enabled in the simulations reported.
		
		When fruiting occurs, fruits are dropped at a random position between an inner and outer calculated radius, with radii satisfying 
		$max(h \cdot r_{in} \cdot 0.5, 0.3)$, 
		where, 
		$h=0.005m_{pp}$ is the height of the PP and 
		$r_{in}=0.2$ is the genetically set inner exclusion ratio.
		
		A genetically determined proportion ($p_{surf}=0.05$) of the seeds produced remain on the surface for a period of 10 s, and then transition below the surface. During their period at the surface seeds can be picked up and stored by Ents (details below). After a genetically determined  `germination' period of duration $t_{pause}=5$ (s), seeds begin to grow symmetrically out from the surface and deeper in to the ground.
		
		\textbf{Growth Condition}  When illuminated, PPs add biomass in increments $\Delta m_{pp}$ while $m_{pp}<m_{ppMax}$. 
		\begin{equation}
			\Delta m_{pp} = E_s \cdot c_{s,m} \cdot m_{c} \cdot \frac{g_{pp}}{\rho_{E_{pp}}} \cdot \Delta t \cdot P
		\end{equation}
		where 
		$E_s=1.0$ is the energy input from the external source, 
		$c_{s,m}=2$ is the external energy collected (collection area $\times$ conversion efficiency), 
		$g_{pp}=0.5$ is a growth constant (kg growth per s, per kg of biomass),  
		$\rho_{E_{pp}}=5$  is the body mass energy density.
		
		\textbf{Competition for space}  To ensure some degree of spacing between PPs, PPs mimic the behaviour of many real-world primary producers and exert direct competitive pressure on their neighbours. A PP competes with another PP if it is within its inner ($r_{inner}$) and outer ($r_{outer}$) competition radii. If they compete, the larger of the two competing PPs has a 10\% chance (per $\Delta t$) of killing the smaller PP. 
		\begin{equation}
			r_{inner} = h \cdot  \zeta_x
		\end{equation}
		
		\begin{equation}
			r_{outer} = h_{max} \cdot \zeta_x
		\end{equation}
		
		where $h_{max} = 0.005 \cdot m_{max_{pp}}=0.25  m$ is the maximum PP height and $ \zeta_x=3$ is the genetically-set exclusion-distance-to-height ratio.
		
		\textbf{Death} PPs die if (i) they reach the genetically-determined maximum age (G[0]=100 s), (ii) they are out-competed (as outlined above), or (iii) they are consumed to the extent their biomass is reduced to zero.
		Upon dying, the PP does not instantaneously disappear, but reduces its remaining mass linearly at a rate $-m_{fade}$.

		%%%%%%%%%%%%%%%%%%%%%%%%%%%%%%%%%%%%%%%%%%%%%%%%%%%%%%%%%%%%%%%%%%%%%

		\subsection{Ents}\label{sec:Ents}
		
		%Two update schedules for Ents: (i) automatic updates (e.g. digestion, growth) which happen each 0.25 s ($\nu=1/0.25$) irrespective of the choices made by Ents, and (ii) actions the Ent chooses to take, which are updated immediately within the same time step. 
		
		\subsubsection{Energy}\label{sec:Energy}
		Energy available to Ents is obtained from consumption and can be stored within their bodies, converted to biomass (growth), used to meet basic metabolic costs and to undertake actions, all of which have an associated energy cost. Energy is quantified in arbitrary `energy units' and referred to inter-changeably as currency-zero ($C_0$).
		
		\textbf{Metabolic Cost} Base metabolic cost is subtracted automatically.
		\begin{equation}
			k_{base} = \nu \cdot b \cdot L^{0.667}
		\end{equation}
		
		where, $k_{base}$ is base metabolic cost rate, $L$ is the lean mass of the entity, $I$ is the ratio of infrequent updates, used to adjust the calculation based on the frequency of updates. %$0.667$ is the exponent applied to the lean mass to account for non-linear scaling effects
		
		\textbf{Consumption} If an Ent is touching an object, it automatically evaluates whether the object is a soft and therefore whether it can be bitten (Ents, PPs and processed blue-cakes are soft). If the Ent chooses to bite the object, it consumes some of its mass, which is immediately removed from the other object. The Ent obtains energy by digesting the consumed mass, the amount of which depends on multiple factors, including the mass, energy density and palatability of the consumed material.
		\begin{equation}
			m_{eaten} = \min(q_{max} - q, m_{obj})
		\end{equation}
		where, $m_{eaten}$ is the mass consumed, $q_{max}$ is the maximum stomach capacity (calculated as $m_{ent} \times q_f$, where $m_{ent}$ is the lean bodymass of the Ent (see below) and $q_f$ is the fractional stomach capacity), $q$ is the current mass in the stomach, and $m_{obj}$ is the mass of the object.
		The mass consumed is digested over time in a series of increments, and as each increment is digested the Ent receives energy.
		\begin{equation}
			\Delta q = \left({q_{max}}/{50t_{dig}}\right)
	 	\end{equation}
		where, $\Delta q$ is the digestion mass increment, $q$ is the stomach content, and $t_{dig}$ is the digestion time needed.
		\begin{equation}
			t_{dig} = q / (q_{max} \cdot d_r)
		\end{equation}
		where, $t_{dig}$ is the digestion time needed, and $d_r$ is the basic rate at which the stomach can process its contents.
		The energy derived from each digested increment is calculated as
		\begin{equation}
			\Delta C_{0_{con}} = \rho_{E_{ent}} \cdot \xi_{con} \cdot \Delta q \cdot \eta
		\end{equation}
		where, $\Delta C_{0_{con}}$ is the energy available, 
		$\rho_{E_{ent}}$ is the energy density of the consumed material (E/kg),
		$\xi_{con}$ is the conversion efficiency for consumed mass, and 
		$\eta$ is the nutritional value coefficient.
		
		For processed blue-cake, the nutritional value is set as $\eta=1$. For other consumed material, $\eta$ depends on the \textit{palatability} of the consumed material, which is determined by `key' $\{K_1 , K_2 , K_3\}$ values of the consumer Ent and the `lock' $\{L_1 , L_2 , L_3\}$ values and of the Ent or PP being consumed.
		
		Palatability ($P$) is based on genetic compatibility over the lock and key values
		\begin{equation}
			P = \frac{\sum_i1 - (K_i + L_i)}{n}
		\end{equation}
		with $n=3$ in this case. The nutrition coefficient is bounded within $[-1,1]$, calculated as
		\begin{equation}
			n_v = \max(\min(1 - 4 \cdot |p|, 1), -1)
		\end{equation}
		where $|\cdot|$ is the absolute value.
		
		\textbf{Energy partitioning} After energy is obtained from digestion, it is routed to either biomass growth or storage (for later use) via a growth-to-free-energy ratio $g$. Ents maintain a reserve of energy, as a buffer against scarcity, and the size of this reserve ($E_{res}$) is controlled by gene value G[34].
		The partitioning ratio $g$ is determined by:
		\begin{equation}
			g = 
			\begin{cases} 
				0 & \text{if } L \geq m_{entMax}, \\
				\max(0, \min(1, g_0 \cdot (1 - m_{ent} \cdot g_{con}))) & \text{if } L < m_{entMax} \text{ and } C_0 < E_{\text{res}}, \\
				1 & \text{if } L < m_{entMax} \text{ and } C_0 \geq E_{\text{res}}.
			\end{cases}
			\label{energyPartitioning}
		\end{equation}
		Here,
		$m_{ent}$ is the lean mass of the Ent,
		$m_{entMax}$ is the maximum lean mass (G[0]), 
		$C_0$ is the amount of energy being stored (as fat) by the Ent, 
		$E_{res}$ is the energy reserve level, 
		$g_0=0.5$ is the initial value of $g$ (G[8]), 
		$g_{con}$ is the growth-energy constant defined by
		\begin{equation}
			g_{con} = g_{\Delta {max}} / m_{entMax} \cdot g_0
		\end{equation}
		where, 
		$g_{\Delta {max}} = 0$ in the present case, meaning the second condition in Eq. \ref{energyPartitioning} reduces to $g_0$.
		
		This logic gives priority to filling the energy reserve; once this full, energy is shared equally between growth and additional storage until growth is complete; then, energy obtained from digestion is all sent to storage for later use.

		\textbf{Growth} 
		Growth occurs incrementally once each 0.25 s. 
		\begin{equation}
			\Delta m_{ent} = \Delta C_{0_{con}} \cdot g / \rho_{E_{ent}} 
		\end{equation}
		where $\rho_{E_{ent}}$ is the energy density (E/kg) of the Ent.
		
		\textbf{Starvation}
		If an Entity does not eat, it continues to operate by using its energy reserve. If its reserve gets depleted, biomass is converted directly to energy (equivalent to catabolism in biological life) at a rate
		
		\begin{equation}
			\Delta C_{0_{con}} = \Delta m_{ent} \cdot \psi
		\end{equation}
		where $\psi=0.9$ (G[47]) is the conversion efficiency between Ent biomass and energy.
		This process can continue until the Ent's body mass ($m_{ent}$) reaches a minimum viable level, after which the Ent dies. This `point of starvation' is dynamically updated, being a fixed fraction of the largest previously-achieved lean biomass.
		\begin{equation}
			m_{starv, t} = \max(m_{starv, t-1}, m_{ent} \cdot f_{starv})
		\end{equation}
		where, $m_{starv, t}$ is the current minimum viable body mass, $m_{starv, t-1}$ is the value until the previous time step, $m_{ent}$ is the lean mass, and $ f_{starv}$ is the minimum viable body mass fraction.

		\subsubsection{Movement}\label{sec:Movement}
		
		Ents can choose to move, which involves deciding on rotation and applied force. The simulation is physically based and so other factors also influence Ent movement such as collision with other (potentially moving) objects, friction (both linear and rotational) acting against the applied forces, and gravity.
		
		In each time step (0.02 $s$), Ents update their rotation by choosing an action $a_{rot}=\{-1,0,1\}$, which defines their change in rotation $\Delta \Phi$.
		\begin{equation}
			\Delta \Phi = \Delta t \cdot t_{rot}^{-1} \cdot a_{rot} \cdot 360
		\end{equation}
		where,
		$t_{rot}$ is the time required for a single rotation and is set be gene G[23], 
		$a_{rot}$ is the chosen rotation action (either of $\{-1,0,1\}$), 
		and the value 360 converts the rotation to degrees.
		
		Similarly, Ents choose how to increment their velocity by applying force $F$ in either a forwards or backwards direction aligned with their rotation angle. 
		\begin{equation}
			F = \max(F_{max} \cdot m_{ent} \cdot (1-q_f), 100)
		\end{equation}
		where $F_{max}$ is the maximum force per kilogram the Ent can apply (set by gene value G[9]), $m_{ent}$ is the lean mass, $q_f$ is the stomach mass ratio. 
		The maximum velocity of Ents ($v_{max}=1$ m/s) is set by gene value G[46]. 
		A minimum value of $F=100$ is enforced so that even small Ents can overcome friction and move. 
		The inclusion of $(1-q_f)$ means there is a trade-off when evolution is enabled between larger stomach size (ability to consume faster) and ability to move quickly. The total mass of the Ent (which the applied force is acting on) includes lean mass, fat mass, stomach content and also any items being carried, and the force in proportion to net lean mass represents the force generating part of the Ent body (muscle, ligaments, bone). 
		
		\textbf{Movement cost} Generating force for movement costs energy and this is calculated as $k_{move}$ based on the magnitude of force applied and a predefined rate of energy consumption.
		\begin{equation}
			k_{move} = v_k \cdot F
		\end{equation}
		where, $v_k=-4\times10^{-5}$ is the energy consumption rate per unit of force. An amusing mistake during the early phase of this work led to the sign of velocity being reversed in the movement cost calculation. During training, Ents quickly learnt they could maximize their energy inputs, without going to the trouble of eating, by travelling backwards as fast as possible in tight circles.
		
		\textbf{Collision damage} Movement can also be associated with a different cost, that associated with collision damage. For some sense of realism, Ents should learn not to collide at high speeds with solid objects. 		
		The impact damage calculation models the collision of two objects based on their kinetic properties.
		Assuming the position vector of the first object is \(\vec{p}_1\) and the point of collision is \(\vec{c}\), the vector \(\vec{d}\), pointing from the position of the first object to the point of collision, is calculated as
		\begin{equation}
			\vec{d} = \vec{c} - \vec{p}_1
		\end{equation}
		The velocity component along the collision direction is given by
		\begin{equation}
			v_{line} = \vec{d} \cdot \vec{v} / \|\vec{d}\|
		\end{equation}	
		where \(\vec{v}\) is the velocity vector of the object, and \(\|\vec{d}\|\) is the magnitude of the collision direction vector. 
		The effective mass used in the damage calculation is
		\begin{equation}
			m_{\text{eff}} = \min(10m_1, m_2)
		\end{equation}
		where \(m_1\) is the mass of the first object and \(m_2\) is the mass of the second object. This formulation limits the impact of extremely disparate mass values on the calculation.
		The relative impact damage is calculated by
		\begin{equation}
			D_{\text{rel}} = -\left(\max(0, v_{\text{line}}) \cdot m_{\text{eff}} \cdot k\right) / (m_1 + m_{\text{eff}})
		\end{equation}
		Here, \(k\) is a coefficient representing the impact damage sensitivity, which scales the damage based on the physical properties and the dynamics of the collision. The use of \(\max(0, v_{\text{line}})\) ensures that only impacts where objects move towards each other are considered, avoiding negative damage values from receding collisions.
		The total impact damage inflicted by the collision is then computed as
		\begin{equation}
			D = D_{\text{rel}} \cdot m_1 \cdot \psi
		\end{equation}
		where $\psi$ is the same conversion efficiency between biomass and energy as described above in the context of starvation.

		\subsubsection{Reproduction and inheritance}\label{sec:Reproduction and inheritance}
		
		Ents do not have different sexes, and each can act as both a `mother' (which gives birth to offspring) and a `father' (which is required to pass on their genes to mothers, triggering reproduction). Ents can enter the `mother' state if the following conditions are satisfied:
		\begin{equation}
			m_{ent} \geq m_\text{repMin} \quad \text{and} \quad C_0 \geq E_\text{birthTotal},
			\label{reproduction_conditions}
		\end{equation}
		where 
		$m_{ent}$ represents the lean mass of the entity; 
		$m_\text{repMin}$ is the minimum mass required for reproduction; 
		$C_0$ is stored energy; 
		$E_\text{birthTotal}$ is the total energy cost required to produce an offspring.
		This total energy cost is met by the mother and is defined
		\begin{equation}
			E_\text{birthTotal} = -(n_\text{offspring} \cdot m_\text{offspring} \cdot \rho_{ent}) - C_{0,0} 
		\end{equation}
		where $n_\text{offspring}$ is the number of offspring produced (fixed at 1 in the present case),  $m_\text{offspring}=m_{entMax} \cdot o_{relMass}$ is the mass of each offspring ($o_{relMass}=G[19]$ being the genetically-determined relative offspring mass), and $C_{0,0}$ is the energy given by the mother which goes in to the offspring's (fat) store (the mother also contributes initial amounts of currency $C_1$ and $C_2$ equal to $C_{initial}=0.1$; G[43]).
		Upon satisfying the conditions of Eq.\ref{reproduction_conditions} the probability of entering the mother state (per 0.25 s) is $p_{\text{rep}}$, where
		\begin{equation}
			p_{\text{rep}} = \nu_{r} \cdot (m_\text{RepMin}/m_{entMax})^{-1} / (\nu_{global} \cdot a_{max} / \lambda_{rMax})
		\end{equation}
		where $\nu_{r}=25$ is the timsteps between each reproductioni state update,  $\nu_{global}=50$ is the global update frequency ($s^{-1}$), $\lambda_{rMax}$ is the genetically-determined expected number of offspring produced per maximum Ent lifetime.
		
		Once a `mother' Ent is in the reproductive state, in order to produce an offspring a `father' Ent must physically touch the `mother'. When this happens, the genetic distance between the two entities ($\Delta_g$) is used to assess their compatibility:
		\begin{equation}
			\Delta_g = \frac{\sum_{i=1}^{3} | g_{1i} - g_{2i} |}{3}
			\label{Delta_g}
		\end{equation}
		where, $g_{1i}$ is the ith gene of the first entity, and \( g_{2i} \) is the ith gene of the second entity.
		Reproduction is successful if $\Delta_g < \Delta_{g_{max}}$, with $\Delta_{g_{max}}=0.15$ being the maximum allowable genetic difference for reproductive compatibility. This value ($\Delta_{g_{max}}$) is itself determined by a gene value, G[42].
		If reproduction is allowed, the offspring's gene values are a random selection of its parents' genes (with each gene value having a 50\% chance of coming  from either parent). Once selected, each value is perturbed by noise, according to
		\begin{equation}
			G[i] = G[i] + \delta_{G[i]} \cdot \tilde{U}
		\end{equation}
		where $\delta_{G[i]}=0.015$ is a `copying noise' parameter, determined by gene value G[56], which means the rate of evolution is to some extent controlled by evolution; $\tilde{U}$ is a uniform random number over the range [-1,1].
		Upon successful reproduction, the Ent offspring appears in the simulated environment next to the mother Ent. It then contributes observations, actions and rewards to the learning process along with other Ents. 

		\subsubsection{Reward updating}
		The RULE algorithm is outlined in the main paper. Constants $\alpha_i$ and $\beta_i$ (where $i$ is reward number) are calculated as follows: run simulation under background conditions with chosen $\theta_i$; while running, record rewards accumulated at each time bin $\tau$ for each Ent, and continue simulation until mean accumulated reward in each age bin are stable, yielding expected values $E_i(\tau)$; set $T_i=\sum_{\tau}E_i(\tau)$,  $n_i=T_i/\theta_i$, $\alpha_i=T_i/n_u$, where $n_u=25$ was user-defined, reflecting the number of generations over which reward adjustment occurs ($n_u$ could of course be added to the gene to make this an evolved choice).
		Recall that $\alpha$ is the increment for updating $E_i(\tau)$ under RULE and $\beta_i$ values increment $\theta_i$. An unstable response (where $\theta_i$ values are always driven to either their minimum or maximum values) is observed when $\beta\geq T_i/n^2$, and so $\beta_i=c_iT_i/n^2$, with $c_i<1$.
		
		\subsubsection{Observations}
		Observations are made automatically (Ents do not have to choose to make observations).
		
		\textbf{`Vision'}
		Ents do not have access to images obtained from their point of view. This approach was tried in the early phases of development, and while it worked, it offered no advantages over the chosen approach, was more computationally costly, and produced slower learning. Therefore, observations analogous to vision are made by Ents using ray-casting within Unity. Notionally, 17 beams are sent from the Ent (a central beam at 0deg, and 8 equally spaced beams either side, providing an 80 deg field of view). Each beam is a cone, expanding away from the Ent to a sphere of radius 0.05 m. The length of each beam $	l_{ray}$ is determined by the strength of the external energy (light) source, so Ents `see' further in stronger light. 
		\begin{equation}
			l_{ray} = max(1, -15+65 \cdot s_{light})
		\end{equation}
		where $s_{light}$ is the light intensity in normalised units, which for convenience is equal to the normalised energy source value $E_s$.
		Each beam returns data on whether the beam hit anything, what that object was (a one-hot encoding of an object tag list), and distance to the object. This information is updated each 0.02 s, and contributes to the total set of observation used to inform Ent behaviour and learning. In addition to this `vision' data, 22 other observations are listed below in Table \ref{tab:observations}.

		\begin{table}[h!]
			\centering
			\begin{tabular}{cll}
				\toprule
				\textbf{Observation} & \textbf{Observation Description} & \textbf{Values} \\
				\midrule
				1 & Normalized X velocity of rigidbody & [0, 1] \\
				2 & Normalized Z velocity of rigidbody & [0, 1]\\
				3 & Normalized Y rotation angle & [0, 1] \\
				4 & Starvation level & [0, 1] \\
				5 & Raw material presence & \{0, 1\} \\
				6 & Processed material presence & \{0, 1\} \\
				7 & Carrying cube & \{0, 1\} \\
				8 & Reproduction readiness & \{0, 1\} \\
				9 & Normalized currency 0 & [0, 1] \\
				10 & Normalized $C_1$ & [0, 1] \\
				11 & Normalized $C_2$ & [0, 1] \\
				12 & Normalized $C_3$ & [0, 1] \\
				13 & Normalized $C_4$ & [0, 1] \\
				14 & Absence of broadcast signal & \{0, 1\} \\
				15 & Broadcast signal 1 detected & \{0, 1\} \\
				16 & Broadcast signal 2 detected & \{0, 1\} \\
				17 & Genetic distance of closest entity & $\geq 0$ \\
				18 & Normalized distance to broadcast signal source & [0, 1] \\
				19 & Normalized angle to broadcast signal source & [0, 1] \\
				20 & Normalised number of entities within detection range & $\geq 0$ \\
				21 & Detect being bitten/eaten & \{-1, 0\} \\
				22 & External energy source strength & $\geq 0$ \\
				\midrule
				& [in addition, Vision information from 17 rays cast] & \\
				\bottomrule
			\end{tabular}
			\caption{Ent observations. \{\} indicate discrete sets of observations, [ ] indicate continuous inclusive ranges. Each Ent collects and processes each observation every 0.02 s. See text for information on vision observations. The term `normalized by' is used to mean the raw value is divided by a normalization factor.
				[1,2] Velocity is normalized by maximum velocity G[46]; 
				[3,19] Rotation angle is normalized by 360; 
				[4]	The Starvation Level is defined $S = 1 - (m_{starv, t} / m_{ent})$; 
				[9-13] All currency values are normalized by 100; 
				[17] Genetic distance to the physically nearest Ent is calculated as $\Delta_g$ (Eq.\ref{Delta_g}), and is referred to as a sense of smell in the main body of the paper; 
				[18] Distance is normalised to the sensing radius (G[41]); 
				[20] The number of Ents within the sensing radius is divided by 100.}
			\label{tab:observations}
		\end{table}
		
		\subsubsection{Actions}\label{sec:Actions}
		
		\begin{table}[h!]
			\centering
			\begin{tabular}{cll}
				\toprule
				\textbf{Action} & \textbf{Action Description} & \textbf{Equation / Condition} \\
				\midrule
				1 & Attempt to bite object &  \{0=False, 1=True\}\\
				2 & Give currency & \{0=False, 1=Give $C_1$, 2=Give $C_2$, 3=Give $C_3$, 4=Give $C_4$\}\\
				3 & Synthesize currency &   \{0=False, 1=True\}\\
				4 & Send signal & \{0=False, 1=Emit\_signal\_1, 2=Emit\_signal\_2\} \\
				5 & Update rotation &  \{-1=Rotate\_aniticlockwise, 0=No\_change, 1=Rotate\_clockwise\}\\
				6 & Update speed &   \{-1=Apply\_negative\_force, 0=No\_change, 1=Apply\_positive\_force\}\\
				7 & Attempt to pick up object & \{0=False, 1=True\}\\
				8 & Drop object &  \{0=False, 1=True\}\\
				\bottomrule
			\end{tabular}
			\caption{Ent discrete action choices. Choices are made by each Ent every 0.02 s.}
			\label{tab:agent_actions}
		\end{table}

		\textbf{Picking-up objects} Entities in a simulated environment have the capability to interact with and pick up objects based on specific conditions. This mechanism is crucial for simulating realistic interactions and behaviors, such as resource gathering or item collection. Even if an entity has the intent to collect objects from its environment, it can only pick up objects lighter than itself, simulating a basic physical constraint.
		\begin{equation}
			m_{object} < m_{entity}
		\end{equation}
		where \(m_{object}\) is the mass of the object being considered for pickup, and \(m_{entity}\) is the mass of the entity.
		
		The entity's storage capacity is equal to its (lean) mass, so new objects can only be stored when: 
		\begin{equation}
			w_{storage} + m_{object} < m_{entity}
		\end{equation}
		where \(w_{storage}\) is the current total weight in the entity's current storage.

		\paragraph{Training}
		The Proximal Policy Optimisation, PPO ~\cite{PPO}, algorithm was employed for RL, implemented through the ML-Agents Toolkit within Unity ~\cite{juliani2020}. A Python API provides a low-level Python interface for communication between the simulation and relevant PyTorch libraries. In the present case, only a single type of Ent is present in the simulations and PPO is implemented with a single policy, with all active agents contributing to the learning process and sharing that same policy. The following hyper-parameters were used during model training: hidden-layers=2x256, gamma=0.99 (discount factor for future rewards), batch-size=32 (number of experiences in each iteration of gradient descent), buffer-size=320000 (the number of experiences collected before updating the policy), time-horizon=32 (the number of steps of experience to collect per-agent before adding to the experience buffer),	learning-rate=0.000025 (gradient descent rate, held constant), beta=0.0035 (strength of the entropy regularization, which ensures exploration by adding randomness to the policy; held constant throughout learning), epsilon=0.15 (a control on the evolution rate of the policy during training - the allowable divergence between current and updated policies; held constant constant during training), lambda=0.99 (a regularisation parameter, weighting the contributions from current and updated policy).
		During the initial training phase, the maximum number of steps was set to 1e8 (recall there are 50 steps per second, and typically approx.100 agents contributing to the process, so this equates to a time limit of approx. 20,000 s). In subsequent experiments the maximum number of steps was set to infinity, and an upper time limit of 25,000 s was enforced.

\begin{table}[h!]
	\begin{tabular}{llllll}
		\toprule
		\textbf{Gene Number} & \textbf{Description} & \textbf{Symbol} & \textbf{Units} & \textbf{Range} & \textbf{Initial Value} \\
		\midrule
		G[0]* & Maximum biomass & $m_{EntMax}$ & kg & [0.01, 50] & 25 \\
		G[1] & Lock L1 \& rgb[0] & $L_1$ & - & [0, 1] & 0.05 \\
		G[2] & Lock L2 \& rgb[1] & $L_2$ & - & [0, 1] & 0.20 \\
		G[3] & Lock L3 \& rgb[2] & $L_3$ & - & [0, 1] & 0.40 \\
		G[4] & Key K1 & $K_1$ & - & [0, 1] & 0.10 \\
		G[5] & Key K2 & $K_2$ & - & [0, 1] & 0.30 \\
		G[6] & Key K3 & $K_3$ & - & [0, 1] & 0.10 \\
		G[7]* & Maximum age & $a_{max}$ & s & [20, 300] & 200 \\
		G[8]* & Initial Growth: Free Energy ratio & $g_0$ & - & [0.2, 0.8] & 0.5 \\
		G[9]* & Maximum Force/Kg & $F_{max}$ & - & [0, 35] & 30 \\
		G[10] & - & -& - & - & - \\
		G[11] & - & -& - & - & - \\
		G[12] & Reward for currency C1 & $\theta_1$ & - & [-1 , 1] & 0.10 \\
		G[13] & Reward for currency C2 & $\theta_2$ & - & [-1 , 1] & 0.10 \\
		G[14] & Reward for currency C3 & $\theta_3$ & - & [-1 , 1] & 0.10 \\
		G[15] & Reward for currency C4 & $\theta_4$ & - & [-1 , 1] & 0.30 \\
		G[16] & Tag number & - & - & $\geq 1$ & 1 \\
		G[17]* & Reproduction min. required mass fraction & $m_\text{repMin}$ & - & [0.2, 1] & 0.75 \\
		G[18] & Offspring per reproduction & $n_\text{offspring}$ & - & $\geq 1$ & 1 \\
		G[19]* & Relative mass of offspring & $o_{relMass}$ & - & [0, 0.5] & 0.2 \\
		G[20] & -& - & - & - & - \\
		G[21] & Reward for consumption & $\theta_5$ & - & [0, 1] & 0.25 \\
		G[22] & Minimum viable body mass fraction & $f_{starv}$ & - & [0.2 - 0.9] & 0.75 \\
		G[23] & Time for single rotation & $t_{rot}$ & s & [0.9 - 1.1] & 1 \\
		G[24] & Digestion rate constant &  $ \xi_{con} $  & $s^{-1}$ & [0, 1] & 0.2 \\
		G[25] & Max change in Growth: Free Energy & $g_{\Delta{max}}$ & - & $\geq 0$  & 0 \\
		G[26] & Reward for reproduction & $\theta_6$ & - & [0, 1] & 1 \\
		G[27]* & Mean reproductions per max lifetime & $\lambda_{rMax}$ & - & $\geq 1$ & 4 \\
		G[28] & Alpha for currency C1 & $\alpha_{C1}$ & - & $\geq0$ & 1.87e-3 \\
		G[29] & Alpha for currency C2 & $\alpha_{C2}$ & - & $\geq0$ & 1.25e-2 \\
		G[30] & Alpha for currency C3 & $\alpha_{C3}$ & - & $\geq0$ & 1.83e-2 \\
		G[31] & Alpha for currency C4 & $\alpha_{C4}$ & - & $\geq0$ & 1.13e-3 \\
		G[32] & Alpha for consumption & $\alpha_{\text{eat}}$ & - & $\geq0$ & 6.2e-1 \\
		G[33] & Alpha for reproduction & $\alpha_{\text{rep}}$ & - & $\geq0$ & 8.29e-1 \\
		G[34]* & Energy reserve (C0) & $C_{0_\text{res}}$ & - & $\geq0$& 1 \\
		G[35] & Synthesizability of C1 & $z_1$ & - & \{0, 1\} & 1 \\
		G[36] & Synthesizability of C2 & $z_2$ & - & \{0, 1\} & 0 \\
		G[37] & Synthesizability of C3 & $z_3$ & - & \{0, 1\} & 1 \\
		G[38] & Synthesizability of C4 & $z_4$ & - & \{0, 1\} & 0 \\
		G[39] & - & - & - & - & - \\
		G[40] & Entity type & - & - & $\geq1$ & 1 \\
		G[41]* & Sensing radius & $r_{sensing}$ & m & $\geq0$ & 3 \\
		G[42]* & Maximum gene difference & $\Delta_{g_{max}}$ & - & [0, 0.2] & 0.15 \\
		G[43]* & Initial currency provided & $C_{initial}$ & - & $\geq0$ & 0.1 \\
		G[44] & Impact damage coefficient & $k$ & - & $\geq0$ & 0.001 \\
		G[45] & Consumption conversion efficiency & $\xi_{con}$ & - & $\geq0$ & 0.9 \\
		G[46]* & Maximum velocity& $v_{max}$ & m/s & [0, 3] & 1 \\
		G[47] & Self mass conversion efficiency & $\psi$ & - & $\geq0$ & 0.9 \\
		G[48] & Movement cost rate & $\nu_k$ & - & $\leq0$ & -4e-5 \\
		G[49] & Transaction cost & $k_{trans}$ & - & $\geq0$ & 0 \\
		G[50] & Currency amount given & $C_{given}$ & - & $\geq0$ & 0.1 \\
		\bottomrule
	\end{tabular}
	\caption{Ent Gene values 0-50.  * indicates the gene values which change under evolution.}
	\label{tab:EntGenes1}
\end{table}

\begin{table}[h!]
	\begin{tabular}{llllll}
		\toprule
		\textbf{Gene Number} & \textbf{Description} & \textbf{Symbol} & \textbf{Units} & \textbf{Range} & \textbf{Initial Value} \\
		\midrule
		G[51] & Base metabolic running cost & $k_{base}$ & E/kg & $\leq0$ & -0.002 \\
		G[52]* & Synthesized currency & $C_{synth}$ & - & 0 - 1 & 0.02 \\
		G[53] & Pain reward & $\theta_{pain}$ & - & [-1, 1] & -1 \\
		G[54] & Number of Locks & $n_{locks}$ & - & $\geq1$ & 3 \\
		G[55] & Energy density & $\rho_{E_{ent}}$ & E/kg & $\geq0$ & 1 \\
		G[56]* & Gene copying noise & $ \delta_{G[i]} $ & - & [0.01, 0.05] & 0.015 \\
		G[57]* & Stomach mass ratio & $q_f$ & - & [0.05, 0.5] & 0.2 \\
		G[58] & C0 given to offspring at birth & $C_{0,0}$ & - & $\geq0$ & 1 \\
		G[59] & - & - & - & - & - \\
		G[60] & Beta value for currency C1 & $\beta_{C1}$ & - & $\geq0$ & 3.8e-3 \\
		G[61] & Beta value for currency C2 & $\beta_{C2}$ & - & $\geq0$ & 3.8e-3 \\
		G[62] & Beta value for currency C3 & $\beta_{C3}$ & - & $\geq0$ & 3.8e-3 \\
		G[63] & Beta value for currency C4 & $\beta_{C4}$ & - & $\geq0$ & 1.14e-2 \\
		G[64] & Beta value for currency consumption & $\beta_{\text{eat}}$ & - & $\geq0$ & 9.5e-3 \\
		G[65] & Beta value for currency reproduce & $\beta_{\text{rep}}$ & - & $\geq0$ & 5e-3\\
		G[66]* & Currency 1 amount given & $C_{1given}$ & - & $\geq0$ & 0.1 \\
		G[67]* & Currency 2 amount given & $C_{2given}$ & - & $\geq0$ & 0.1 \\
		G[68]* & Currency 3 amount given & $C_{3given}$ & - & $\geq0$ & 0.1 \\
		G[69] & Currency 4 amount given & $C_{4given}$ & - & $\geq0$ & 1 \\
		\bottomrule
	\end{tabular}
	\caption{Ent Gene values 51-69. * indicates the gene values which change under evolution.}
	\label{tab:EntGenes2}
\end{table}

\end{document}